\documentclass[10pt,twocolumn,letterpaper]{article}

\usepackage{iccv}
\usepackage{times}
\usepackage{epsfig}
\usepackage{graphicx}
\usepackage{amsmath}
\usepackage{amssymb}
\usepackage{mathtools}
\usepackage{mynotation}
\usepackage{booktabs}
\usepackage{algorithm}
\usepackage{algpseudocode}

\usepackage{bbm}
\usepackage{color}
\usepackage{stfloats}

\usepackage[caption=false]{subfig}

\algrenewcommand{\algorithmicrequire}{\textbf{Input:}}
\algrenewcommand{\algorithmicensure}{\textbf{Output:}}

\usepackage[pagebackref=true,breaklinks=true,letterpaper=true,colorlinks,bookmarks=false]{hyperref}

\iccvfinalcopy 

\def\iccvPaperID{3941} 
\def\httilde{\mbox{\tt\raisebox{-.5ex}{\symbol{126}}}}

\ificcvfinal\fi
\begin{document}

\title{Learning Discriminative Model Prediction for Tracking}

\newcommand{\asep}{\hspace{6mm}}
\newcommand{\aand}{\hspace{6mm}}
\author{Goutam Bhat$^{*}$  \aand  Martin Danelljan$^{*}$ \aand Luc Van Gool \aand Radu Timofte \vspace{1.5mm}\\
	CVL, ETH Z\"urich, Switzerland
}

\maketitle

\newcommand{\parsection}[1]{\noindent\textbf{#1:} }

\begin{abstract}
	The current strive towards end-to-end trainable computer vision systems imposes major challenges for the task of visual tracking. In contrast to most other vision problems, tracking requires the learning of a robust target-specific appearance model \emph{online}, during the inference stage. To be end-to-end trainable, the online learning of the target model thus needs to be embedded in the tracking architecture itself. Due to the imposed challenges, the popular Siamese paradigm simply predicts a target feature template, while ignoring the background appearance information during inference. Consequently,  the predicted model possesses limited target-background discriminability.

We develop an end-to-end tracking architecture, capable of fully exploiting both target and background appearance information for target model prediction. Our architecture is derived from a discriminative learning loss by designing a dedicated optimization process that is capable of predicting a powerful model in only a few iterations. Furthermore, our approach is able to learn key aspects of the discriminative loss itself. The proposed tracker sets a new state-of-the-art on 6 tracking benchmarks, achieving an EAO score of $0.440$ on VOT2018, while running at over $40$ FPS. The code and models are available at \url{https://github.com/visionml/pytracking}.%
{\let\thefootnote\relax\footnote{{$^*$Both authors contributed equally.}}}
\end{abstract}


\section{Introduction}
Generic object tracking is the task of estimating the state of an arbitrary target in each frame of a video sequence. In the most general setting, the target is only defined by its initial state in the sequence. Most current approaches address the tracking problem by constructing a target model, capable of differentiating between the target and background appearance. Since target-specific information is only available at test-time, the target model cannot be learned in an offline training phase, as in for instance object detection. Instead, the target model must be constructed during the inference stage itself by exploiting the target information given at test-time. 
This unconventional nature of the visual tracking problem imposes significant challenges when pursuing an end-to-end learning solution.

The aforementioned problems have been most successfully addressed by the Siamese learning paradigm~\cite{SiameseFC,SiamRPN}.
These approaches first learn a feature embedding, where the similarity between two image regions is computed by a simple cross-correlation. Tracking is then performed by finding the image region most similar to the target template. In this setting, the target model simply corresponds to the template features extracted from the target region. Consequently, the tracker can easily be trained end-to-end using pairs of annotated images.

\begin{figure}[!t]
	\centering%
	\newcommand{\wid}{0.33\columnwidth}%
	\includegraphics*[trim = 0 70 0 60, width = \wid]{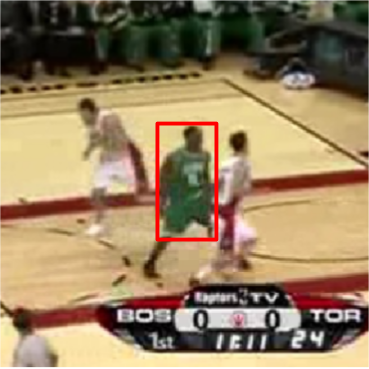}%
	\includegraphics*[trim = 0 70 0 73, width = \wid]{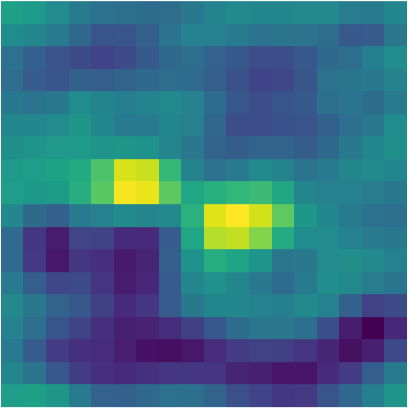}%
	\includegraphics*[trim = 0 70 0 73, width = \wid]{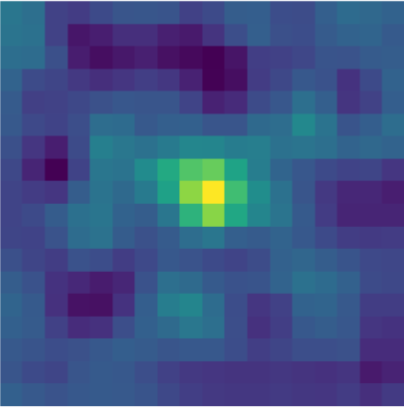}
	\includegraphics*[trim = 0 60 20 110, width = \wid]{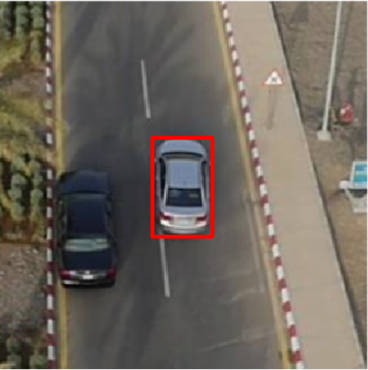}%
	\includegraphics*[trim = 0 60 20 125, width = \wid]{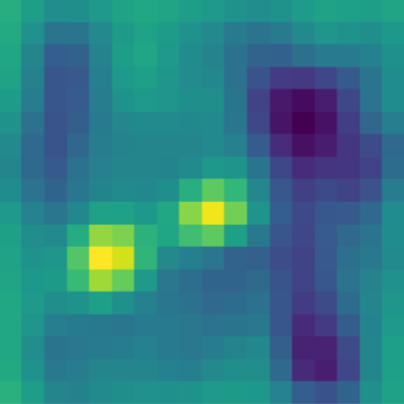}%
	\includegraphics*[trim = 0 60 20 125, width = \wid]{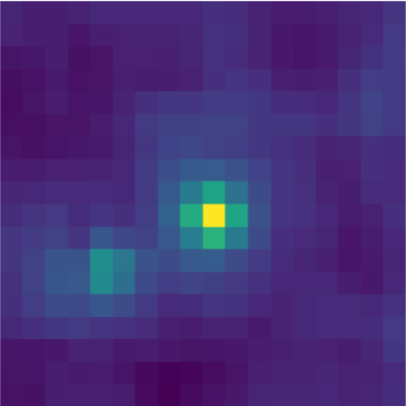}
	\includegraphics*[trim = 0 100 40 80, width = \wid]{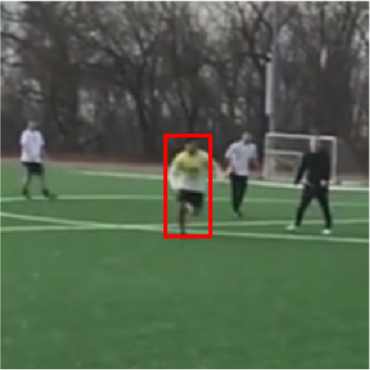}%
	\includegraphics*[trim = 0 100 40 95, width = \wid]{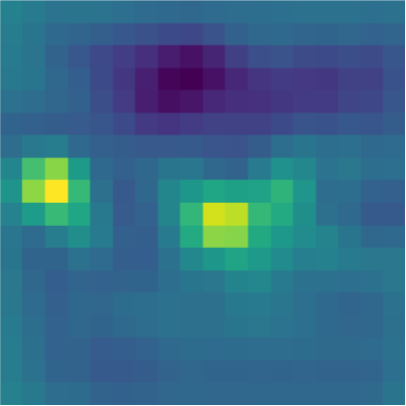}%
	\includegraphics*[trim = 0 100 40 95, width = \wid]{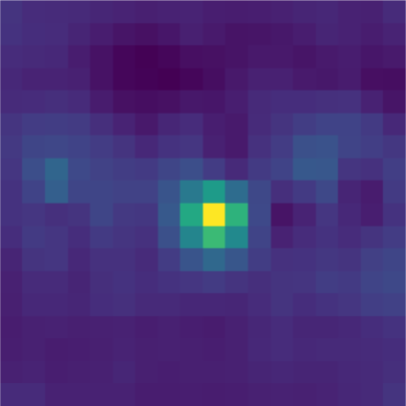}\vspace{-0.5mm}
	\parbox{.32\columnwidth}{\centering\small Image}%
	\parbox{.32\columnwidth}{\centering\small Siamese based}%
	\parbox{.32\columnwidth}{\centering\small Ours}
	\caption{
		Confidence maps of the target object (red box) provided by the target model obtained using i) a Siamese approach (middle), and ii) Our  approach (right). The model predicted in a Siamese fashion, using only target appearance, struggles to distinguish the target from distractor objects in the background. In contrast, our model prediction architecture also integrates background appearance, providing superior discriminative power.}\vspace{-2.5mm}%
	\label{fig:intro}%
\end{figure}

Despite its recent success, the Siamese learning framework suffers from severe limitations. Firstly, Siamese trackers only utilize the target appearance when inferring the model. This completely ignores background appearance information, which is crucial for discriminating the target from similar objects in the scene (see figure~\ref{fig:intro}). Secondly, the learned similarity measure is not necessarily reliable for objects that are not included in the offline training set, leading to poor generalization. Thirdly, the Siamese formulation does not provide a powerful model update strategy. Instead, state-of-the-art approaches resort to simple template averaging \cite{DaSiamRPN}. These limitations result in inferior robustness \cite{VOT2018} compared to other state-of-the-art tracking approaches. 

In this work, we introduce an alternative tracking architecture, trained in an end-to-end manner, that directly addresses all aforementioned limitations. In our design, we take inspiration from the \emph{discriminative} online learning procedures that have been successfully applied in recent trackers \cite{ATOM,DanelljanECCV2016,MDNet}. Our approach is based on a target model prediction network, which is derived from a discriminative learning loss by applying an iterative optimization procedure. The architecture is carefully designed to enable effective end-to-end training, while maximizing the discriminative ability of the predicted model. This is achieved by ensuring a minimal number of optimization steps through two key design choices. First, we employ a steepest descent based methodology that computes an optimal step length in each iteration. Second, we integrate a module that effectively initializes the target model. Furthermore, we introduce significant flexibility into our final architecture by learning the discriminative learning loss itself.

Our entire tracking architecture, along with the backbone feature extractor, is trained using annotated tracking sequences by minimizing the prediction error on future frames. We perform comprehensive experiments on 7 tracking benchmarks: VOT2018~\cite{VOT2018}, LaSOT~\cite{LaSOT}, TrackingNet~\cite{TrackingNet}, GOT10k~\cite{GOT10k}, NFS~\cite{NfS}, OTB-100~\cite{OTB2015}, and UAV123~\cite{UAV123}. Our approach achieves state-of-the-art results on all 7 datasets, while running at over 40 FPS. We also provide an extensive experimental analysis of the proposed architecture, showing the impact of each component.

\section{Related Work}
Generic object tracking has undergone astonishing progress in recent years, with the development of a variety of approaches. Recently, methods based on Siamese networks~\cite{SiameseFC,SiamRPN,Tao2016Sint} have received much attention due to their end-to-end training capabilities and high efficiency. The name derives from the deployment of a Siamese network architecture in order to learn a similarity metric \emph{offline}. Bertinetto \etal~\cite{SiameseFC} utilize a fully-convolutional architecture for similarity prediction, thereby attaining high tracking speeds of over 100 FPS. Wang \etal~\cite{RASNet} learn a residual attention mechanism to adapt the tracking model to the current target. Li \etal~\cite{SiamRPN} employ a region proposal network~\cite{FasterRCNN} to obtain accurate bounding boxes.

A key limitation in Siamese approaches is their inability to incorporate information from the background region or previous tracked frames into the model prediction. A few recent attempts aim to address these issues. Guo \etal~\cite{DSiam} learn a feature transformation to handle the target appearance changes and to suppress background. Zhu \etal~\cite{DaSiamRPN} handle background distractors by subtracting corresponding image features from the target template during online tracking. Despite these attempts, the Siamese trackers are yet to reach high level of robustness attained by state-of-the-art trackers employing online learning \cite{VOT2018}. 

In contrast to Siamese methods, another family of trackers \cite{ATOM,DanelljanCVPR2017,MDNet} learn a discriminative classifier \emph{online} to distinguish the target object from the background. These approaches can effectively utilize background information, thereby achieving impressive robustness on multiple tracking benchmarks~\cite{VOT2018,OTB2015}. However, such methods rely on more complicated online learning procedures that cannot be easily formulated in an end-to-end learning framework. Thus, these approaches are often restricted to features extracted from deep networks pre-trained for image classification \cite{DanelljanECCV2016,HCF_ICCV15} or hand-crafted alternatives \cite{DanelljanICCV2015}. 

A few recent works aim to formulate existing discriminative online learning based trackers as a neural network component in order to benefit from end-to-end training. Valmadre \etal~\cite{Valmadre2017cvpr} integrate the single-sample closed-form solution of the correlation filter (CF)~\cite{Henriques14} into a deep network. Yao \etal~\cite{RTINet} unroll the ADMM iterations in BACF~\cite{BACFgaloogahi} tracker to learn the feature extractor and a few tracking hyper-parameters in a complex multi-stage training procedure. The BACF model learning is however restricted to the single-sample variant of the Fourier-domain CF formulation which cannot exploit multiple samples, requiring ad-hoc linear combination of filters for model adaption. 

The problem of learning to predict a target model using only a few images is closely related to meta-learning~\cite{Finn2017MAML,Munkhdalai2017MetaN,Naik1992MetaneuralNT,Ravi2017OptimizationAA,schmidhuber1987srl,Schmidhuber1992LCF,Thrun1998Learningtolearn}. A few works have already pursued this direction for tracking. Bertinetto \etal~\cite{Bertinetto2016LearningFO} meta-train a network to predict the parameters of the tracking model.  
Choi \etal~\cite{DeepMetaLearning} utilize a meta-learner to predict a target-specific feature space to complement the general target-independent feature space used for estimating the similarity in Siamese trackers. 
Park \etal~\cite{MetaTracker} develop a meta-learning framework employing an initial target independent model, which is then refined using gradient descent with learned step-lengths.
However, constant step-lengths are only suitable for fast initial adaption of the model and does not provide optimal convergence when applied iteratively. 

\section{Method}

\begin{figure*}[t]
	\centering%
	\newcommand{\wid}{0.75\textwidth}%
	\includegraphics*[trim = 0 0 0 0, width = \wid]{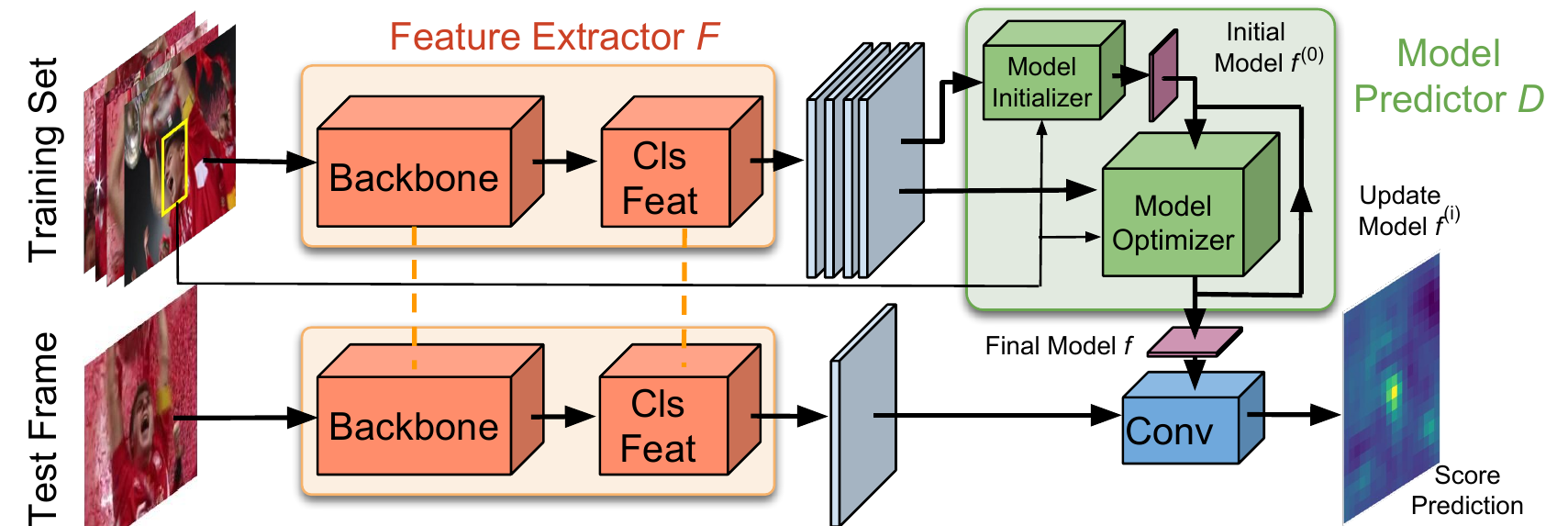}%
	\caption{An overview of the target classification branch in our tracking architecture. Given an annotated training set (top left), we extract deep feature maps using a backbone network followed by an additional convolutional block (Cls Feat). The feature maps are then input to the model predictor $D$, consisting of the initializer and the recurrent optimizer module. The model predictor outputs the weights of the convolutional layer which performs target classification on the feature map extracted from the test frame.}%
	\label{fig:method}%
	\vspace{-3mm}
\end{figure*}

In this work, we develop a discriminative model prediction architecture for tracking. As in Siamese trackers, our approach benefits from end-to-end training. However, unlike Siamese, our architecture can fully exploit background information and provides natural and powerful means of updating the target model with new data. Our model prediction network is derived from two main principles: (i) A discriminative learning loss promoting robustness in the learned target model; and (ii) a powerful optimization strategy ensuring rapid convergence. By such careful design, our architecture can predict the target model in only a few iterations, without compromising its discriminative power.

In our framework, the target model constitutes the weights of a convolutional layer, providing target classification scores as output. Our model prediction architecture computes these weights by taking a set of bounding-box annotated image samples as input. The model predictor includes an initializer network that efficiently provides an initial estimate of the model weights, using only the target appearance. These weights are then processed by the optimizer module, taking both target and background appearance into account. By design, our optimizer module possesses few learnable parameters in order to avoid overfitting to certain classes and scenes during offline training. Our model predictor can thus generalize to unseen objects, which is crucial in \emph{generic} object tracking. 

Our final tracking architecture consists of two branches: a target classification branch (see figure~\ref{fig:method}) for distinguishing the target from background, and a bounding box estimation branch for predicting an accurate target box. 
Both branches input deep features from a common backbone network. The target classification branch contains a convolutional block, extracting features on which the classifier operates. Given a training set of samples and corresponding target boxes, the model predictor generates the weights of the target classifier. These weights are then applied to features extracted from the test frame, in order to compute the target confidence scores. For the bounding box estimation branch, we utilize the overlap maximization based architecture introduced in \cite{ATOM}. The entire tracking network, including the target classification, bounding box estimation and backbone modules, is trained offline on tracking datasets.

\subsection{Discriminative Learning Loss}
\label{sec:disc_learn_loss}

In this section, we describe the discriminative learning loss used to derive our model prediction architecture. The input to our model predictor $D$ consists of a training set $S_\text{train}=\{(x_j, c_j)\}_{j=1}^n$ of deep feature maps $x_j \in \mathcal{X}$ generated by the feature extractor network $F$. Each sample is paired with the corresponding target center coordinate $c_j \in \reals^2$. Given this data, our aim is to predict a target model $f = D(S_\text{train})$. The model $f$ is defined as the filter weights of a convolutional layer tasked with discriminating between target and background appearance in the feature space $\mathcal{X}$. We gather inspiration from the least-squares-based regression take on the tracking problem, that has seen tremendous success in the recent years \cite{ATOM,DanelljanCVPR2017,Henriques14}. However, in this work we generalize the conventional least-squares loss applied for tracking in several directions, allowing the final tracking network to \emph{learn} the optimal loss from data.

In general, we consider a loss of the form,
\begin{equation}
\label{eq:trainloss}
L(f) = \frac{1}{|S_\text{train}|} \sum_{(x,c) \in S_\text{train}} \|r(x \conv f, c)\|^2 + \|\lambda f\|^2 \,. 
\end{equation}
Here, $\conv$ denotes convolution and $\lambda$ is a regularization factor. The function $r(s, c)$ computes the residual at every spatial location based on the target confidence scores $s = x \conv f$ and the ground-truth target center coordinate $c$. The most common choice is $r(s, c) = s - y_c$, where $y_c$ are the desired target scores at each location, popularly set to a Gaussian function centered at $c$ \cite{MOSSE2010}. However, simply taking the difference forces the model to regress calibrated confidence scores, usually zero, for all negative samples. This requires substantial model capacity, forcing the learning to focus on the negative data samples instead of achieving the best discriminative abilities. Furthermore, taking the na\"ive difference does not address the problem of data imbalance between target and background.

To alleviate the latter issue of data imbalance, we use a spatial weight function $v_c$. The subscript $c$ indicates the dependence on the center location of the target, as detailed in section~\ref{sec:learning_loss}. To accommodate the first issue, we modify the loss following the philosophy of Support Vector Machines. We employ a hinge-like loss in $r$, clipping the scores at zero as $\max(0, s)$ in the background region. The model is thus free to predict large negative values for easy samples in the background without increasing the loss. For the target region on the other hand, we found it disadvantageous to add an analogous hinge loss $\max(0, 1 - s)$. Although contradictory at a first glance, this behavior can be attributed to the fundamental asymmetry between the target and background class, partially due to the numerical imbalance. Moreover, accurately calibrated target confidences are indeed advantageous in the tracking scenario, e.g.\ for detecting target loss. We therefore desire the properties of standard least-squares regression in the target neighborhood.

To accommodate the advantages of both least-squares regression and the hinge loss, we define the residual function,
\begin{equation}
\label{eq:resfunc}
r(s, c) = v_c \cdot \left(m_c s + (1 - m_c) \max (0, s) - y_c \right) \,.
\end{equation}
The target region is defined by the mask $m_c$, having values in the interval $m_c(t) \in [0,1]$ at each spatial location $t \in \reals^2$. Again, the subscript $c$ indicate the dependence on the target center coordinate. The formulation in \eqref{eq:resfunc} is capable of continuously changing the behavior of the loss from standard least squares regression to a hinge loss depending on the image location relative to the target center $c$. Setting $m_c \approx 1$ at the target and $m_c \approx 0$ in the background region yields the desired behavior described above. However, how to optimally set $m_c$ is not clear, in particular at the transition region between target and background. While the classical strategy is to manually set the mask parameters using trial and error, our end-to-end formulation allows us to learn the mask in a data-driven manner. In fact, as detailed in section~\ref{sec:learning_loss}, our approach learns all free parameters in the loss: the target mask $m_c$, the spatial weight $v_c$, the regularization factor $\lambda$, and even the regression target $y_c$ itself.

\subsection{Optimization-Based Architecture}
\label{sec:optim_arch}

Here, we derive the network architecture $D$ that predicts the filter $f = D(S_\text{train})$ by implicitly minimizing the error \eqref{eq:trainloss}. The network is designed by formulating an optimization procedure. From eqs.~\eqref{eq:trainloss} and \eqref{eq:resfunc} we can easily derive a closed-form expression for the gradient of the loss $\nabla L$ with respect to the filter $f$\footnote{See supplementary material (section~\ref{sec:grad_loss}) for details.}. The straight-forward option is to then employ gradient descent using a step length $\alpha$,
\begin{equation}
\label{eq:gd}
f^{(i+1)} = f^{(i)} - \alpha \nabla L(f^{(i)}) \,.
\end{equation}
However, we found this simple approach to be insufficient, even if the learning rate $\alpha$ (either a scalar or coefficient-specific) is learned by the network itself (see section~\ref{sec:exp_baseline}). It experiences slow adaption of the filter parameters $f$, requiring a vast increase in the number of iterations. This harms efficiency and complicates offline learning.

The slow convergence of gradient descent is largely due to the constant step length $\alpha$, which does not depend on data or the current model estimate. We solve this issue by deriving a more elaborate optimization approach, requiring only a handful of iterations to predict a strong discriminative filter $f$. The core idea is to compute the step length $\alpha$ based on the steepest descent methodology, which is a common optimization technique \cite{NumericalOptimization,CGpain}. We first approximate the loss with a quadratic function at the current estimate $f^{(i)}$,
\begin{align}
\label{eq:lossquad}
L(f) \approx \tilde{L}(f) = & \frac{1}{2} (f - f^{(i)})\tp Q^{(i)} (f - f^{(i)}) + \\ \nonumber
& (f - f^{(i)})\tp \nabla L(f^{(i)}) + L(f^{(i)}) \,.
\end{align}
Here, the filter variables $f$ and $f^{(i)}$ are seen as vectors and $Q^{(i)}$ is positive definite square matrix. The steepest descent then proceeds by finding the step length $\alpha$ that minimizes the approximate loss \eqref{eq:lossquad} in the gradient direction \eqref{eq:gd}. This is found by solving $\frac{\diff}{\diff \alpha} \tilde{L}\left(f^{(i)} - \alpha \nabla L(f^{(i)})\right) = 0$, as
\begin{equation}
\label{eq:steplength}
\alpha = \frac{\nabla L(f^{(i)})\tp \nabla L(f^{(i)})}{\nabla L(f^{(i)})\tp Q^{(i)} \nabla L(f^{(i)})} \,.
\end{equation}
In steepest descent, the formula \eqref{eq:steplength} is used to compute the scalar step length $\alpha$ in each iteration of the filter update \eqref{eq:gd}.

The quadratic model \eqref{eq:lossquad}, and consequently the resulting step length \eqref{eq:steplength}, depends on the choice of $Q^{(i)}$. For example, by using a scaled identity matrix $Q^{(i)} = \frac{1}{\beta} I$ we retrieve the standard gradient descent algorithm with a fixed step length $\alpha = \beta$. On the other hand, we can now integrate second order information into the optimization procedure. The most obvious choice is setting $Q^{(i)} = \frac{\partial^2 L}{\partial f^2}(f^{(i)})$ to the Hessian of the loss \eqref{eq:trainloss}, which corresponds to a second order Taylor approximation \eqref{eq:lossquad}. For our least-squares formulation \eqref{eq:trainloss} however, the Gauss-Newton method \cite{NumericalOptimization} provides a powerful alternative, with significant computational benefits since it only involves first-order derivatives. We thus set $Q^{(i)} = (J^{(i)})\tp J^{(i)}$, where $J^{(i)}$ is the Jacobian of the residuals at $f^{(i)}$. In fact, neither the matrix $Q^{(i)}$ or Jacobian $J^{(i)}$ need to be constructed explicitly, but rather implemented as a sequence of neural network operations. See the supplementary material (section~\ref{sec:h}) for details. Algorithm~\ref{alg:model-predictor} describes our target model predictor $D$. Note that our optimizer module can easily be employed for online model adaption as well. This is achieved by continuously extending the training set $S_\text{train}$ with new samples from the previously tracked frames. The optimizer module is then applied on this extended training set, using the current target model as the initialization $f^{(0)}$.

\subsection{Initial Filter Prediction}
\label{sec:initializer}
To further reduce the number of optimization recursions required in $D$, we introduce a small network module that predicts an initial model estimate $f^{(0)}$. Our initializer network consists of a convolutional layer followed by a precise ROI pooling~\cite{IOUNet}. The latter extracts features from the target region and pools them to the same size as the target model $f$. The pooled feature maps are then averaged over all the samples in $S_\text{train}$ to obtain the initial model $f^{(0)}$. As in Siamese trackers, this approach only utilizes the target appearance. However, rather than predicting the final model, our initializer network is tasked with only providing a reasonable initial estimate, which is then processed by the optimizer module to provide the final model.

\newcommand{\init}{\mathtt{ModelInit}}
\newcommand{\filtergrad}{\mathtt{FiltGrad}}
\newcommand{\assign}{\leftarrow}
\newcommand{\algcomment}[2]{\hspace{#2mm}{\footnotesize\# #1}}
\begin{algorithm}[t]
	\caption{Target model predictor $D$.}
	\begin{algorithmic}[1]
		\Require Samples $S_\text{train}=\{(x_j, c_j)\}_{j=1}^n$, iterations $N_\text{iter}$
		\State $f^{(0)} \assign \init(S_\text{train})$ \algcomment{Initialize filter (sec~\ref{sec:initializer})}{8}
		\For{$i = 0, \ldots, N_\text{iter} - 1$} \algcomment{Optimizer module loop}{8}
		\State $\nabla L(f^{(i)}) \assign \filtergrad(f^{(i)}, S_\text{train})$ \algcomment{Using \eqref{eq:trainloss}-\eqref{eq:resfunc}}{3}
		\State $h \assign J^{(i)} \nabla L(f^{(i)})$ \algcomment{Apply Jacobian of \eqref{eq:resfunc}}{14}
		\State $\alpha \assign \| \nabla L(f^{(i)})\|^2 / \|h\|^2$ \algcomment{Compute step length \eqref{eq:steplength}}{5.5}
		\State $f^{(i+1)} \assign f^{(i)} - \alpha \nabla L(f^{(i)})$ \algcomment{Update filter}{5}
		\EndFor
	\end{algorithmic}
	\label{alg:model-predictor}
\end{algorithm}

\subsection{Learning the Discriminative Learning Loss}
\label{sec:learning_loss}
Here, we describe how the free parameters in the residual function~\eqref{eq:resfunc}, defining the loss \eqref{eq:trainloss}, are learned. Our residual function includes the label confidence scores $y_c$, the spatial weight function $v_c$ and the target mask $m_c$. While such variables are constructed by hand in current discriminative online learning based trackers, our approach in fact learns these functions from data. We parametrize them based on the distance from the target center. This is motivated by the radial symmetry of the problem, where the direction to the sample location relative to the target is of little significance. In contrast, the distance to the sample location plays a crucial role, especially in the transition from target to background. Thus, we parameterize $y_c$, $m_c$ and $v_c$ using radial basis functions $\rho_k$ and learn their coefficients $\phi_k$. For instance, the label $y_c$ at position $t \in \reals^2$ is given by
\begin{equation}
\label{eq:masks_parametrization}
y_c(t) = \sum_{k=0}^{N-1} \phi_k^y \rho_k(\|t - c\|)\,.
\end{equation}

We use triangular basis functions $\rho_k$, defined as
\begin{equation}
\label{eq:rbf}
\rho_k(d)= 
\begin{cases}
\max(0, 1 - \frac{|d - k \Delta|}{\Delta}), & k < N-1\\
\max(0, \min(1, 1 + \frac{d - k\Delta}{\Delta})),  & k = N - 1
\end{cases}
\end{equation}
The above formulation corresponds to a continuous piecewise linear function with a knot displacement of $\Delta$. Note that the final case $k=N-1$ represents all locations that are far away from the target center and thus can be treated identically. We use a small $\Delta$ to enable accurate representation of the regression label at the target-background transition. The functions $v_c$ and $m_c$ are parameterized analogously using coefficients $\phi_k^v$ and $\phi_k^m$ respectively in \eqref{eq:masks_parametrization}. For the target mask $m_c$, we constrain the values to the interval $[0,1]$ by passing the output from \eqref{eq:masks_parametrization} through a Sigmoid function. 

\begin{figure}[t]
	\centering%
	\newcommand{\wid}{0.6\columnwidth}%
	\includegraphics*[trim = 0 0 0 0, width = \wid]{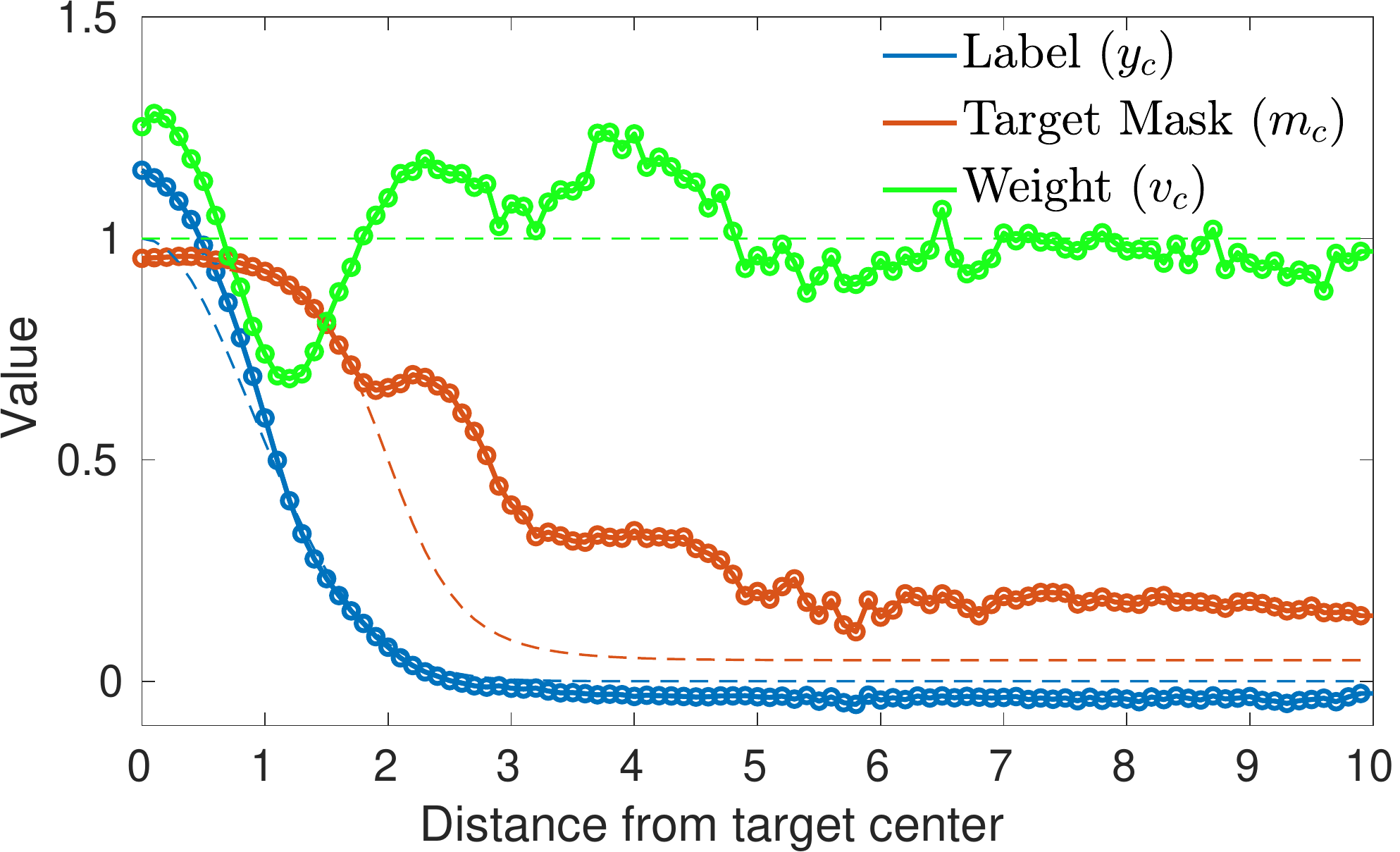}\vspace{0mm}%
	\caption{Plot of the learned regression label ($y_c$), target mask ($m_c$), and spatial weight ($v_c$). The markers show the knot locations. The initialization of each quantity is shown in dotted lines.}%
	\label{fig:learned_loss_shrunk}\vspace{-3mm}%
\end{figure}

We use $N=100$ basis functions and set the knot displacement to $\Delta=0.1$ in the resolution of the deep feature space $\mathcal{X}$. For offline training, the regression label $y_c$ is initialized to the same Gaussian $z_c$ used in the offline classification loss, described in section~\ref{sec:offline_training}. The weight function $v_c$ is initialized to constant $v_c(t) = 1$. Lastly, we initialize the target mask $m_c$ using a scaled $\tanh$ function. The coefficients $\phi_k$, along with $\lambda$, are learned as part of the model prediction network $D$ (see section~\ref{sec:offline_training}). The initial and learned values for $y_c$, $m_c$ and $v_c$ are visualized in figure~\ref{fig:learned_loss_shrunk}. Notably, our network learns to increase the weight $v_c$ at the target center and reduce it in the ambiguous transition region.

\subsection{Bounding Box Estimation}
We utilize the overlap maximization strategy introduced in~\cite{ATOM} for the task of accurate bounding box estimation. 
Given a reference target appearance, the bounding box estimation branch is trained to predict the IoU overlap between the target and a set of 
candidate boxes on a test image. The target information is integrated into the IoU prediction by computing a modulation vector from the reference appearance of the target. The computed vector is used to modulate the features from the test image, which are then used for IoU prediction. The IoU prediction network is differentiable w.r.t.\ the input box co-ordinates, allowing the candidates to be refined during tracking by maximizing the predicted IoU. We use the same network architecture as in~\cite{ATOM}.

\subsection{Offline Training}
\label{sec:offline_training}
Here, we describe our offline training procedure. In Siamese approaches, the network is trained with image pairs, using one image to predict the target template and the other for evaluating the tracker. In contrast, our model prediction network $D$ inputs a \emph{set} $S_\text{train}$ of multiple data samples from the sequence. To better exploit this advantage, we train our full tracking architecture on pairs of sets $(M_\text{train}, M_\text{test})$. Each set $M=\{(I_j,b_j)\}_{j=1}^{N_\text{frames}}$ consists of images $I_j$ paired with their corresponding target bounding boxes $b_j$. The target model is predicted using $M_\text{train}$ and then evaluated on the test frames $M_\text{test}$. Uniquely, our training allows the model predictor $D$ to learn how to better utilize multiple samples. The sets are constructed by sampling a random segment of length $T_\text{ss}$ in the sequence. We then construct $M_\text{train}$ and  $M_\text{test}$ by sampling $N_\text{frames}$ frames each from the first and second halves of the segment respectively.

Given the pair $(M_\text{train}, M_\text{test})$, we first pass the images through the backbone feature extractor to construct the train $S_\text{train}$ and test $S_\text{test}$ samples for our target model. Formally, the train set is obtained as $S_\text{train} = \{(F(I_j), c_j): (I_j,b_j) \in M_\text{train}\}$, where $c_j$ is the center coordinate of the box $b_j$. This is input to the target predictor $f = D(S_\text{train})$. The aim is to predict a model $f$ that is discriminative and that generalizes well to future unseen frames. We therefore only evaluate the predicted model $f$ on the test samples $S_\text{test}$, obtained analogously using $M_\text{test}$. Following the discussion in section~\ref{sec:disc_learn_loss}, we compute the regression errors using a hinge for the background samples,
\begin{equation}
	\label{eq:offline-error}
	\ell(s, z) = \begin{dcases}
		s - z \,, & z > T \\
		\max(0, s) \,, & z \leq T
	\end{dcases} \,.
\end{equation}
Here, the threshold $T$ defines the target and background region based on the label confidence value $z$. For the target region $z > T$ we take the difference between the predicted confidence score $s$ and the label $z$, while we only penalize positive confidence values for the background $z \leq T$.

The total target classification loss is computed as the mean squared error \eqref{eq:offline-error} over all test samples. However, instead of only evaluating the final target model $f$, we average the loss over the estimates $f^{(i)}$ obtained in each iteration $i$ by the optimizer (see alg.~\ref{alg:model-predictor}). This introduces intermediate supervision to the target prediction module, benefiting training convergence. Furthermore, we do not aim to train for a specific number of recursions, but rather be free to set the desired number of optimization recursions online. It is thus natural to evaluate each iterate $f^{(i)}$ equally. The target classification loss used for offline training is given by,
\begin{equation}
\label{eq:classification-loss}
L_\text{cls} = \frac{1}{N_\text{iter}} \sum_{i=0}^{N_\text{iter}} \sum_{(x,c) \in S_\text{test}} \left\|\ell\big(x \conv f^{(i)}, z_c\big)\right\|^2 \,.
\end{equation}
Here, regression label $z_c$ is set to a Gaussian function centered as the target $c$. Note that the output $f^{(0)}$ from the filter initializer (section~\ref{sec:initializer}) is also included in the above loss. Although not denoted explicitly to avoid clutter, both $x$ and $f^{(i)}$ in \eqref{eq:classification-loss} depend on the parameters of the feature extraction network $F$. The model iterates $f^{(i)}$ additionally depend on the parameters in the model predictor network $D$. 

For bounding box estimation, we extend the training procedure in \cite{ATOM} to image sets by computing the modulation vector on the first frame in $M_\text{train}$ and sampling candidate boxes from all images in $M_\text{test}$. The bounding box estimation loss $L_\text{bb}$ is computed as the mean squared error between the predicted IoU overlaps in $M_\text{test}$ and the ground truth. We train the full tracking architecture by combining this with the target classification loss \eqref{eq:classification-loss} as $L_\text{tot} = \beta L_\text{cls} + L_\text{bb}$.

\noindent\textbf{Training details:}
We use the training splits of the TrackingNet~\cite{TrackingNet}, LaSOT~\cite{LaSOT}, GOT10k~\cite{GOT10k} and COCO~\cite{COCO} datasets. The backbone network is initialized with the ImageNet weights. We train for 50 epochs by sampling 20,000 videos per epoch, giving a total training time of less than 24 hours on a single Nvidia TITAN X GPU. We use ADAM~\cite{ADAM} with learning rate decay of $0.2$ every 15th epoch. The target classification loss weight is set to $\beta = 10^2$ and we use $N_\text{iter} = 5$ optimizer module recursions in \eqref{eq:classification-loss} during training.
The image patches in $(M_\text{train}, M_\text{test})$ are extracted by sampling a random translation and scale relative to the target annotation. We set the base scale to 5 times the target size to incorporate significant background information. For each sequence, we sample $N_\text{frames} = 3$ test and train frames, using a segment length of $T_\text{ss} = 60$. The label scores $z_c$ are constructed using a standard deviation of $1/4$ relative to the base target size, and we use $T=0.05$ for the regression error \eqref{eq:offline-error}. We employ the ResNet architecture for the backbone. For the model predictor $D$, we use features extracted from the third block, having a spatial stride of 16. We set the kernel size of the target model $f$ to $4 \times 4$. 

\subsection{Online Tracking}
\label{sec:online_tracking}
Given the first frame with annotation, we employ data augmentation strategies \cite{BhatECCV2018} to construct an initial set $S_\text{train}$ containing $15$ samples. The target model is then obtained using our discriminative model prediction architecture $f = D(S_\text{train})$. For the first frame, we employ $10$ steepest descent recursions, after the initializer module. Our approach allows the target model to be easily updated by adding a new training sample to $S_\text{train}$ whenever the target is predicted with sufficient confidence. We ensure a maximum memory size of 50 by discarding the oldest sample. During tracking, we refine the target model $f$ by performing two optimizer recursions every 20 frames, or a single recursion whenever a distractor peak is detected. Bounding box estimation is performed using the same settings as in~\cite{ATOM}.

\section{Experiments}

Our approach is implemented in Python using PyTorch, and operates at 57 FPS with a ResNet-18 backbone and 43 FPS with ResNet-50 on a single Nvidia GTX 1080 GPU. Detailed results are provided in the supplementary material (section~\ref{sec:VOT2018}--\ref{sec:less-data}).

\subsection{Analysis of our Approach}
\label{sec:exp_baseline}
Here, we perform an extensive analysis of the proposed model prediction architecture. Experiments are performed on a combined dataset containing the entire OTB-100 \cite{OTB2015}, NFS (30 FPS version) \cite{NfS} and UAV123 \cite{UAV123} datasets. This pooled dataset contains $323$ diverse videos to enable thorough analysis. The trackers are evaluated using the AUC \cite{OTB2015} metric. Due to the stochastic nature of the tracker, we always report the average AUC score over $5$ runs. We employ ResNet-18 as the backbone network for this analysis.

\begin{table}[!t]
	\centering\vspace{-1mm}
	\resizebox{0.5\columnwidth}{!}{%
		\begin{tabular}{lccc}
\toprule
&Init&GD&SD\\\midrule
AUC&58.2&61.6&63.8\\\bottomrule
\end{tabular}

	}\vspace{1mm}%
	\caption{
		Analysis of different model prediction architectures on the combined OTB-100, NFS and UAV123 datasets. The architecture using only the target information for model prediction (\textbf{Init}) achieves an AUC score of $58.2\%$. The proposed steepest descent based architecture (\textbf{SD}) provides the best results, outperforming the gradient descent method  (\textbf{GD}) by over $2.2\%$ AUC score.}
	\label{tab:ablation_discr_learn}%
	\vspace{-2mm}
\end{table}

\parsection{Impact of optimizer module}
We compare our proposed method, utilizing the steepest descent (\textbf{SD}) based architecture, with two alternative approaches. \textbf{Init:} Here, we only use the initializer module to predict the final target model, which corresponds to removing the optimizer module in our approach. Thus, similar to the Siamese approaches, only target appearance information is used for model prediction, while background information is discarded. \textbf{GD:} In this approach, we replace steepest descent with the gradient descent (GD) algorithm using learned coefficient-wise step-lengths $\alpha$ in \eqref{eq:gd}. All networks are trained using the same settings. The results for this analysis are shown in table~\ref{tab:ablation_discr_learn}. 

The model predicted by the initializer network, which uses only target information, achieves an AUC score of $58.2\%$. The gradient descent approach, which can exploit background information, provides a substantial improvement, achieving an AUC score of $61.6\%$. This highlights the importance of employing discriminative learning for model prediction. Our steepest descent approach obtains the best results, outperforming GD by $2.2\%$. This is due to the superior convergence properties of steepest descent, important for offline learning and fast online tracking.

\begin{table}[!t]
	\centering\vspace{-1mm}
	\resizebox{0.6\columnwidth}{!}{%
		\begin{tabular}{l@{~}c@{~~}c@{~~}c@{~~}c@{~~}c@{~~}}
\toprule
&SD&+Init&+FT&+Cls&+Loss\\\midrule
AUC&58.7&60.0&62.6&63.3&63.8\\\bottomrule

\end{tabular}

%
%

%
	}\vspace{1mm}%
	\caption{
		Analysis of the impact of initializer module (\textbf{+Init}), training the backbone (\textbf{+FT}), using extra conv. block (\textbf{+Cls}) and offline learning of the loss (\textbf{+Loss}), by incrementally adding them one at a time. The baseline \textbf{SD} constitutes our steepest descent based optimizer module along with a ResNet-18 trained on ImageNet.
	}
	\label{tab:ablation_e2e_learn}%
	\vspace{-4mm}
\end{table}

\parsection{Analysis of model prediction architecture} Here, we analyze the impact of key aspects of the proposed discriminative online learning architecture, by incrementally adding them one at a time. The results are shown in table~\ref{tab:ablation_e2e_learn}. The baseline \textbf{SD} constitutes our steepest descent based optimizer module along with a fixed ResNet-18 network trained on ImageNet. That is, similar to the current state-of-the-art discriminative approaches, we do not fine-tune the backbone. Instead of learning the discriminative loss, we employ the regression error \eqref{eq:offline-error} in the optimizer module. This baseline approach achieves an AUC score of $58.7\%$. By adding the model initializer module (\textbf{+Init}), we achieve a significant gain of $1.3\%$ in AUC score. Further training the entire network, including backbone feature extractor, (\textbf{+FT}) leads to a major improvement of $2.6\%$ in AUC score. This demonstrates the advantages of learning specialized features suitable for tracking through end-to-end learning. Using an additional convolutional block to extract classification specific features (\textbf{+Cls}) yields a further improvement of $0.7\%$ AUC score. Finally, learning the discriminative loss \eqref{eq:resfunc} itself (\textbf{+Loss}), as described in section~\ref{sec:learning_loss}, improves the AUC score by another $0.5\%$. This shows the benefit of learning the implicit online loss by maximizing the generalization capabilities of the model on future frames.

\begin{table}[!t]
	\centering\vspace{-1mm}
	\resizebox{0.7\columnwidth}{!}{%
		\begin{tabular}{lccc}
\toprule
&No update&Model averaging&Ours\\\midrule
AUC&61.7&61.7&63.8\\\bottomrule
\end{tabular}

	}\vspace{1mm}%
	\caption{Comparison of different model update strategies on the combined OTB-100, NFS and UAV123 datasets.
	}
	\label{tab:online_update}%
	\vspace{-1mm}
\end{table}

\parsection{Impact of online model update}
Here, we analyze the impact of updating the target model online, using information from previous tracked frames.
We compare three different model update strategies. i) \textbf{No update:} The model is not updated during tracking. Instead, the model predicted in the first frame by our model predictor $D$, is employed for the entire sequence. ii) \textbf{Model averaging:} In each frame, the target model is updated using the linear combination of the current and newly predicted model, as commonly employed in tracking \cite{Henriques14,BACFgaloogahi,Valmadre2017cvpr}. iii) \textbf{Ours:} The target model is obtained using the training set constructed online, as described in section~\ref{sec:online_tracking}. The na\"ive model averaging fails to improve over the baseline method with no updates (see table~\ref{tab:online_update}). In contrast, our approach obtains a significant gain of about $2\%$ in AUC score over both methods, indicating that our approach can effectively adapt the target model online.
\subsection{State-of-the-art Comparison}
We compare our proposed approach \textbf{DiMP} with the state-of-the-art methods on seven challenging tracking benchmarks. Results for two versions of our approach are shown: DiMP-18 and DiMP-50 employing ResNet-18 and ResNet-50 respectively as the backbone network.

\begin{table}[!t]
	\centering\vspace{-1mm}
	\resizebox{1.01\columnwidth}{!}{%
		\begin{tabular}{l@{~}c@{~~}c@{~~}c@{~~}c@{~~}c@{~~}c@{~~}c@{~~}c@{~~}c@{~~}c@{~~}}
	\toprule
	&DRT&RCO&UPDT&DaSiam-&MFT&LADCF&ATOM&SiamRPN++&\textbf{DiMP-18}&\textbf{DiMP-50}\\
	&\cite{DRT}&\cite{VOT2018}&\cite{BhatECCV2018}&RPN \cite{DaSiamRPN}&\cite{VOT2018}&\cite{LADCF}&\cite{ATOM}&\cite{SiamRPN++}&&\\\midrule
	EAO&0.356&0.376&0.378&0.383&0.385&0.389&0.401&\textbf{\textcolor{blue}{0.414}}&0.402&\textbf{\textcolor{red}{0.440}}\\
	Robustness&0.201&0.155&0.184&0.276&\textbf{\textcolor{red}{0.140}}&0.159&0.204&0.234&0.182&\textbf{\textcolor{blue}{0.153}}\\
	Accuracy&0.519&0.507&0.536&0.586&0.505&0.503&0.590&\textbf{\textcolor{red}{0.600}}&0.594&\textbf{\textcolor{blue}{0.597}}\\\bottomrule
\end{tabular}

	}\vspace{1mm}%
	\caption{State-of-the-art comparison on the VOT2018 dataset in terms of expected average overlap (EAO), accuracy \& robustness.}
	\label{tab:vot}%
	\vspace{-4mm}
\end{table}

\parsection{VOT2018 \cite{VOT2018}} We evaluate our approach on the 2018 version of the Visual Object Tracking (VOT) challenge consisting of $60$ challenging videos. Trackers are evaluated using the measures accuracy (average overlap over successfully tracked frames) and robustness (failure rate). Both these measures are combined to get the EAO (Expected Average Overlap) score used to rank trackers. The results are shown in table~\ref{tab:vot}. Among previous approaches, SiamRPN++ achieves the best accuracy and EAO. However, it attains much inferior robustness compared to the discriminative learning based approaches, such as MFT and LADCF. Similar to the aforementioned approaches, SiamRPN++ employs ResNet-50 for feature extraction. Our approach DiMP-50, employing the same backbone network, significantly outperforms SiamRPN++ with a relative gain of $6.3\%$ in terms of EAO. Further, compared to SiamRPN++, our approach has a $34\%$ lower failure rate, while achieving similar accuracy. 
This shows that discriminative model prediction is crucial for robust tracking.

\begin{figure}[t]
	\centering%
	\newcommand{\wid}{0.55\columnwidth}%
	\includegraphics*[trim = 0 0 0 0, width = \wid]{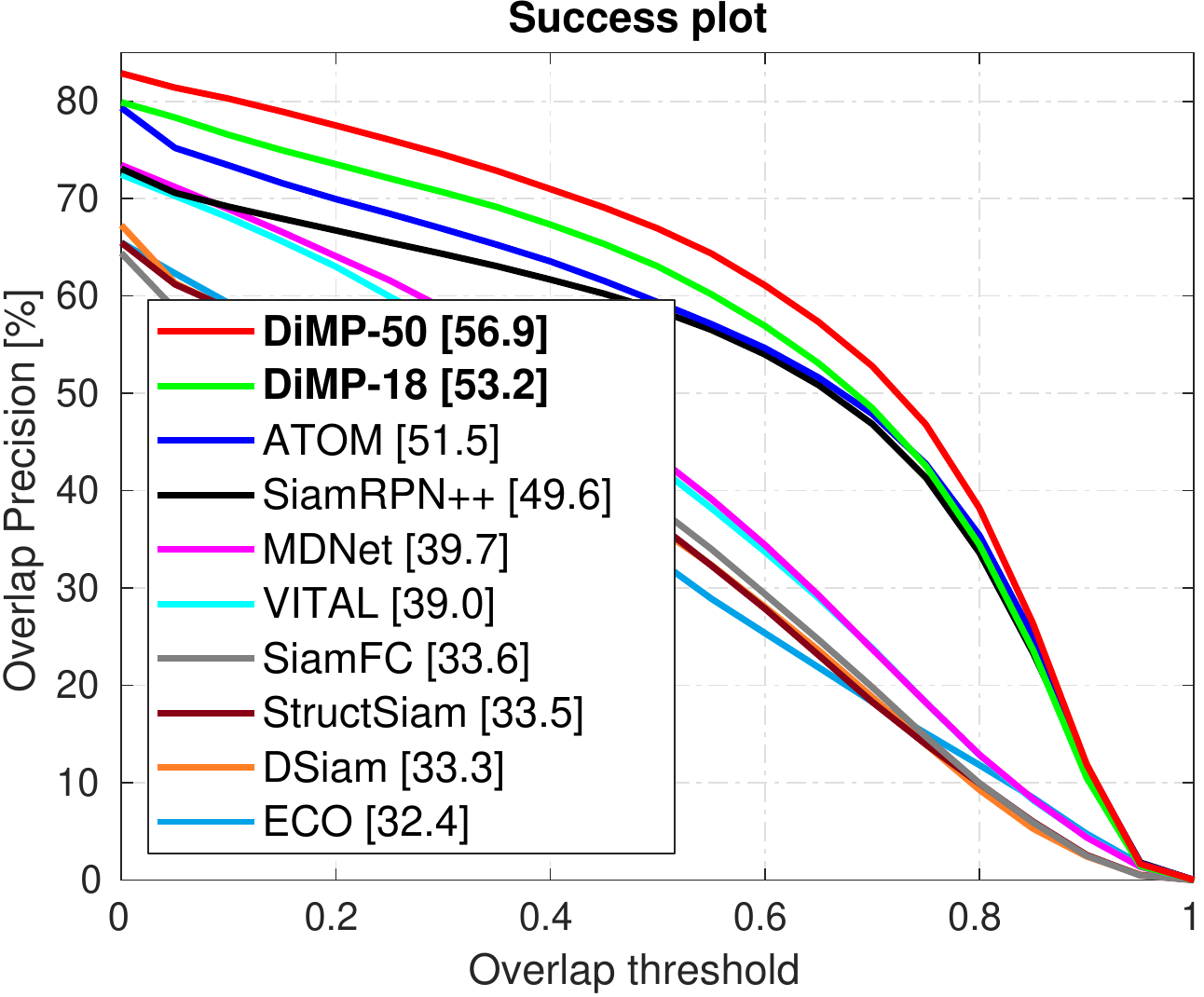}\vspace{-1.5mm}%
	\caption{Success plot on the LaSOT dataset.}%
	\label{fig:lasot}\vspace{-4mm}%
\end{figure}

\parsection{LaSOT \cite{LaSOT}} We evaluate our approach on the test set consisting of $280$ videos. The success plots are shown in figure~\ref{fig:lasot}. Compared to other datasets, LaSOT has longer sequences, with an average of $2500$ frames per sequence. Thus, online model adaption is crucial for this dataset. The previous best approach ATOM \cite{ATOM} employs online discriminative learning with with pre-trained ResNet-18 features. 
Our end-to-end trained approach, using the same backbone architecture, outperforms ATOM with a relative gain of $3.3\%$, showing the impact of end-to-end training. DiMP-50 further improves the results with an AUC score of $56.9\%$. These results demonstrate the powerful model adaption capabilities of our method on long sequences.

\parsection{TrackingNet \cite{TrackingNet}} We evaluate our approach on the test set of the large-scale TrackingNet dataset. The results are shown in table~\ref{tab:trackingnet}. SiamRPN++ achieves an impressive AUC score of $73.3\%$. Our approach, with the same ResNet-50 backbone as in SiamRPN++, outperforms all previous methods by achieving AUC score of $74.0\%$.

\parsection{GOT10k \cite{GOT10k}} This is large-scale dataset containing over $10,000$ videos, $180$ of which form the test set used for evaluation. Interestingly, there is no overlap in object classes between the train and test splits, promoting the importance of generalization to \emph{unseen} object classes. To ensure fair evaluation, the trackers are forbidden from using external datasets for training. We follow this protocol by retraining our trackers using only the GOT10k train split. Results are shown in table~\ref{tab:got10k}. ATOM achieves an average overlap (AO) score of $55.6\%$. Our ResNet-18 version outperforms ATOM with a relative gain of $4.1\%$. 
Our ResNet-50 version achieves the best AO score of $61.1\%$, verifying the strong generalization abilities of our tracker.

\begin{table}[!t]
	\centering\vspace{-1mm}
	\resizebox{1.01\columnwidth}{!}{%
		\begin{tabular}{l@{~}c@{~~}c@{~~}c@{~~}c@{~~}c@{~~}c@{~~}c@{~~}c@{~~}c@{~~}c@{~~}}
	\toprule
	&ECO&SiamFC&CFNet&MDNet&UPDT&DaSiam-&ATOM&SiamRPN++&\textbf{DiMP-18}&\textbf{DiMP-50}\\
	&\cite{DanelljanCVPR2017}&\cite{SiameseFC}&\cite{Valmadre2017cvpr}&\cite{MDNet}&\cite{BhatECCV2018}&RPN \cite{DaSiamRPN}&\cite{ATOM}&\cite{SiamRPN++}&&\\\midrule
	Precision (\%)&49.2&53.3&53.3&56.5&55.7&59.1&64.8&\textbf{\textcolor{red}{69.4}}&66.6&\textbf{\textcolor{blue}{68.7}}\\
	Norm.\ Prec.\ (\%)&61.8&66.6&65.4&70.5&70.2&73.3&77.1&\textbf{\textcolor{blue}{80.0}}&78.5&\textbf{\textcolor{red}{80.1}}\\
	Success (AUC) (\%)&55.4&57.1&57.8&60.6&61.1&63.8&70.3&\textbf{\textcolor{blue}{73.3}}&72.3&\textbf{\textcolor{red}{74.0}}\\\bottomrule
\end{tabular}

	}\vspace{1mm}%
	\caption{State-of-the-art comparison on the TrackingNet test set in terms of precision, normalized precision, and success.}
	\label{tab:trackingnet}%
	\vspace{-2mm}
\end{table}
\begin{table}[!t]
	\centering\vspace{-1mm}
	\resizebox{1.01\columnwidth}{!}{%
		\begin{tabular}{l@{~}c@{~~}c@{~~}c@{~~}c@{~~}c@{~~}c@{~~}c@{~~}c@{~~}c@{~~}c@{~~}}
	\toprule
	&MDNet&CF2&ECO&CCOT&GOTURN&SiamFC&SiamFCv2&ATOM&\textbf{DiMP-18}&\textbf{DiMP-50}\\
	&\cite{MDNet}&\cite{HCF_ICCV15}&\cite{DanelljanCVPR2017}&\cite{DanelljanECCV2016}&\cite{Held2016gotrun}&\cite{SiameseFC}&\cite{Valmadre2017cvpr}&\cite{ATOM}&&\\\midrule
	SR$_{0.50}$ (\%)&30.3&29.7&30.9&32.8&37.5&35.3&40.4&63.4&\textbf{\textcolor{blue}{67.2}}&\textbf{\textcolor{red}{71.7}}\\
	SR$_{0.75}$ (\%)&9.9&8.8&11.1&10.7&12.4&9.8&14.4&40.2&\textbf{\textcolor{blue}{44.6}}&\textbf{\textcolor{red}{49.2}}\\
	AO (\%)&29.9&31.5&31.6&32.5&34.7&34.8&37.4&55.6&\textbf{\textcolor{blue}{57.9}}&\textbf{\textcolor{red}{61.1}}\\\bottomrule
\end{tabular}

	}\vspace{1mm}%
	\caption{State-of-the-art comparison on the GOT10k test set in terms of average overlap (AO), and success rates (SR) at overlap thresholds $0.5$ and $0.75$.}
	\label{tab:got10k}%
	\vspace{-1mm}
\end{table}
\begin{table}[!t]
	\centering\vspace{-1mm}
	\resizebox{1.01\columnwidth}{!}{%
		\begin{tabular}{l@{~}c@{~~}c@{~~}c@{~~}c@{~~}c@{~~}c@{~~}c@{~~}c@{~~}c@{~~}c@{~~}}
	\toprule
	&ECOhc&DaSiam-&ATOM&CCOT&MDNet&ECO&SiamRPN++&UPDT&\textbf{DiMP-18}&\textbf{DiMP-50}\\
	&\cite{DanelljanCVPR2017}&RPN \cite{DaSiamRPN}&\cite{ATOM}&\cite{DanelljanECCV2016}& \cite{MDNet}&\cite{DanelljanCVPR2017}&\cite{SiamRPN++}&\cite{BhatECCV2018}&&\\\midrule
	NFS&-&-&58.4&48.8&41.9&46.6&-&53.6&\textbf{\textcolor{blue}{61.0}}&\textbf{\textcolor{red}{61.9}}\\
	OTB-100&64.3&65.8&66.3&68.2&67.8&69.1&\textbf{\textcolor{blue}{69.6}}&\textbf{\textcolor{red}{70.4}}&66.0&68.4\\
	UAV123&51.2&57.7&64.2&51.3&-&53.2&-&54.5&\textbf{\textcolor{blue}{64.3}}&\textbf{\textcolor{red}{65.3}}\\\bottomrule
\end{tabular}

	}\vspace{1mm}%
	\caption{State-of-the-art comparison on the NFS, OTB-100 and UAV123 datasets in terms of AUC score.}
	\label{tab:nfs_uav_otb}%
	\vspace{-2mm}
\end{table}

\parsection{Need for Speed \cite{NfS}}
We evaluate our approach on the $30$ FPS version of the dataset, containing challenging videos with fast-moving objects. The AUC scores over all the $100$ videos are shown in table~\ref{tab:nfs_uav_otb}. The previous best method ATOM achieves an AUC score of $58.4\%$ . Our approach outperforms ATOM with relative gains of $4.4\%$ and $6.0\%$ using ResNet-18 and ResNet-50 respectively.

\parsection{OTB-100 \cite{OTB2015}} Table~\ref{tab:nfs_uav_otb} shows the AUC scores over all the $100$ videos in the dataset. Among the compared methods, UPDT achieves the best results with an AUC score of $70.4\%$. Our DiMP-50 achieves an AUC score of $68.4\%$, competitive with the other state-of-the-art approaches. 

\parsection{UAV123 \cite{UAV123}} This dataset consists of $123$ low altitude aerial videos captured from a UAV. Results in terms of AUC are shown in table~\ref{tab:nfs_uav_otb}. Among previous methods, ATOM achieves an AUC score of $64.2\%$. Both DiMP-18 and DiMP-50 outperform ATOM, achieving AUC scores of $64.3\%$ and $65.4\%$, respectively.

\section{Conclusions}
We propose a tracking architecture that is trained offline in an end-to-end manner. Our approach is derived from a discriminative learning loss by applying an iterative optimization procedure. By employing a steepest descent based optimizer and an effective model initializer, our approach can predict a powerful model in only a few optimization steps. Further, our approach learns the discriminative loss during offline training by minimizing the prediction error on unseen test frames. Our approach sets a new state-of-the-art on 6 tracking benchmarks, while operating at over $40$ FPS.

\noindent\textbf{Acknowledgments}:
This work was supported by ETH General Fund (OK), and Nvidia through a hardware grant.

{\small
\bibliographystyle{ieee}
\bibliography{references}
}

\clearpage
\newcommand{\comment}[2]{\textcolor{red}{\textbf{#1}: \textit{#2}}}

\newcommand{\charfunc}{\mathbbm{1}}

\def\httilde{\mbox{\tt\raisebox{-.5ex}{\symbol{126}}}}

\setcounter{equation}{0}
\setcounter{figure}{0}
\setcounter{table}{0}
\setcounter{section}{0}

\renewcommand{\theequation}{S\arabic{equation}}
\renewcommand{\thefigure}{S\arabic{figure}}
\renewcommand{\thetable}{S\arabic{table}}
\renewcommand{\thesection}{S\arabic{section}}

\begin{center}
	\textbf{\large Supplementary Material}
\end{center}

This supplementary material provides additional details and results. Section~\ref{sec:grad_loss} derives the closed form expression of the filter gradient, employed in the optimizer module. In section~\ref{sec:h} we derive the application of the Jacobian in order to compute the quantity $h$, employed in algorithm~1 in the paper. In section~\ref{sec:VOT2018} we provide detailed results on the VOT2018 dataset, while in section~\ref{sec:LaSOT}, we provide detailed results on the LaSOT dataset. We also provide additional details on the NFS, OTB100 and UAV123 datasets in section~\ref{sec:detailed-res}. We analyze the impact when training with less data in section~\ref{sec:less-data}. Finally, we provide a 2d visualization of the learned functions parametrizing the discriminative loss in section \ref{sec:label_vis}.

\section{Closed-Form Expression for $\nabla L$}
\label{sec:grad_loss}

Here, we derive a closed-form expression for the gradient of the loss \eqref{eq:trainloss} in the main paper, also restated here,
\begin{equation}
\label{eq:trainloss_s}
L(f) = \frac{1}{|S_\text{train}|} \sum_{(x,c) \in S_\text{train}} \|r(s, c)\|^2 + \|\lambda f\|^2 \,. 
\end{equation}
Here, $s = x \conv f$ is the score map obtained after convolving the deep feature map $x$ with the target model $f$. The training set is given by $S_\text{train} = \{(x_j, c_j)\}_{j=1}^n$. The residual function $r(s,c)$ is defined as (also eq.~\eqref{eq:resfunc} in the paper),
\begin{equation}
\label{eq:resfunc_s}
r(s, c) = v_c \cdot \left(m_c s + (1 - m_c) \max (0, s) - y_c \right) \,.
\end{equation}
The gradient $\nabla L(f)$ of the loss \eqref{eq:trainloss_s} w.r.t.\ the filter coefficients $f$ is then computed as,
\begin{equation}
\label{eq:nalba_l_s}
\nabla L(f) = \frac{2}{|S_\text{train}|} \sum_{(x,c) \in S_\text{train}}  \left(\frac{\partial r_{s,c}}{\partial f}\right)  \tp r_{s,c} + 2\lambda ^2 f \,.
\end{equation}
Here, we have defined $r_{s,c} = r(s, c)$ and $\frac{\partial r_{s,c}}{\partial f}$ corresponds to the Jacobian of the residual function \eqref{eq:resfunc_s} w.r.t.\ the filter coefficients $f$. Using eq.~\eqref{eq:resfunc_s} we obtain, 
\begin{align}
\frac{\partial r_{s,c}}{\partial f} & = \diag(v_c m_c) \frac{\partial s}{\partial f} + \diag\left((1 - m_c) \cdot \charfunc_{s > 0}\right) \frac{\partial s}{\partial f} \nonumber \\
& = \diag(q_c) \frac{\partial s}{\partial f} \,.
\label{eq:jacobian_s}
\end{align}
Here, $\diag(q_c)$ denotes a diagonal matrix containing the elements in $q_c$. Further, $q_c = v_c m_c + (1 - m_c) \cdot \charfunc_{s > 0}$ is computed using only point-wise operations, where $\charfunc_{s > 0}$ is $1$ for positive $s$ and $0$ otherwise. Using eqs.~\eqref{eq:nalba_l_s} and \eqref{eq:jacobian_s} we finally obtain,
\begin{equation}
\label{eq:grad_loss_final_s}
\nabla L(f) = \frac{2}{|S_\text{train}|} \sum_{(x,c) \in S_\text{train}} \!\! \left(\frac{\partial s}{\partial f}\right) \tp \left(q_c \cdot r_{s,c}\right) + 2\lambda ^2 f \,.
\end{equation}
Here, $\cdot$ denotes the element-wise product. The multiplication with the transposed Jacobian $\big(\frac{\partial s}{\partial f}\big) \tp$ corresponds to backpropagation of the input $q_c \cdot r_{s,c}$ through the convolution layer $f \mapsto x \conv f$. This is implemented as a transposed convolution with $x$. The closed-form expression \eqref{eq:grad_loss_final_s} is thus easily implemented using standard operations in a deep learning library like PyTorch.

\section{Calculation of $h$ in Algorithm \ref{alg:model-predictor}}
\label{sec:h}

In this section, we show the calculation of $h = J^{(i)} \nabla L(f^{(i)})$, used when determining the optimal step length $\alpha$ in Algorithm \ref{alg:model-predictor} in the main paper. Since we only need the squared $L^2$ norm of $h$ in step length calculation, we will directly derive an expression for $\|h\|^2 = \|J^{(i)} \nabla L(f^{(i)})\|^2$. Here, $J^{(i)} = \left.\frac{\partial \xi}{\partial f}\right|_{f^{(i)}}$ is the Jacobian of the residual vector $\xi$ of loss \eqref{eq:trainloss_s}, evaluated at the filter estimate $f^{(i)}$. Not to be confused with the residual function \eqref{eq:resfunc_s}, the residual vector $\xi$ is obtained as the concatenation of individual residuals $\xi_j = r(x_j \conv f, c_j) / \sqrt{n} $ for $j \in \{1,\ldots,n\}$ and $\xi_j = \lambda f$ for $j = n + 1$. Here, $n = |S_\text{train}|$ is the number of samples in $S_\text{train}$. Consequently, we get,
\begin{align}
\label{eq:norm_h_s}
& \|h\|^2 =\left\|J^{(i)} \nabla L(f^{(i)})\right\|^2 \\ 
& =  \sum_{j=1}^{n+1} \left\|\left.\frac{\partial \xi_j}{\partial f}\right|_{f^{(i)}} \nabla L(f^{(i)})\right\|^2 \nonumber \\
&=\!\!\sum_{j=1}^{n}  \left\|\frac{1}{\sqrt{n}}\!\left.\frac{\partial r(x_j \conv f, c_j)}{\partial f}\right|_{f^{(i)}} \!\!\!\! \nabla L(f^{(i)})\right\|^2 \!\! + \left\|\lambda \nabla L(f^{(i)})\right\|^2 \nonumber \\
& = \frac{1}{n} \sum_{(x,c) \in S_\text{train}} \left\|\left.\frac{\partial r_{s, c}}{\partial f}\right|_{f^{(i)}} \!\!\! \nabla L(f^{(i)})\right\|^2 + \left\|\lambda \nabla L(f^{(i)})\right\|^2 \,. \nonumber
\end{align} 

Using eqs.~\eqref{eq:norm_h_s} and \eqref{eq:jacobian_s} we finally obtain,
\begin{align}
\begin{split}
\|h\|^2 &= \frac{1}{|S_\text{train}|} \sum_{(x,c) \in S_\text{train}} \left\|q_c \cdot \left(\left.\frac{\partial s}{\partial f}\right|_{f^{(i)}} \nabla L(f^{(i)})\right)\right\|^2 + \\
&\qquad \|\lambda \nabla L(f^{(i)})\|^2 \nonumber \\
&= \frac{1}{|S_\text{train}|} \sum_{(x,c) \in S_\text{train}} \left\|q_c \cdot \left(x \conv \nabla L(f^{(i)})\right)\right\|^2 + \\
&\qquad \|\lambda \nabla L(f^{(i)})\|^2
\end{split}
\end{align}
As described in section \ref{sec:grad_loss}, $\nabla L(f^{(i)})$ is computed using the closed-form expression \eqref{eq:grad_loss_final_s}. The term $\left.\frac{\partial s}{\partial f}\right|_{f^{(i)}} \nabla L(f^{(i)})$ corresponds to convolution of $x$ with $\nabla L(f^{(i)})$, i.e.\ $\left.\frac{\partial s}{\partial f}\right|_{f^{(i)}} \nabla L(f^{(i)}) = x \conv \nabla L(f^{(i)})$. Thus, $\|h\|^2$ is computed easily using standard operations from deep learning libraries.

\begin{figure}[t]
	\centering%
	\newcommand{\wid}{0.95\columnwidth}%
	\includegraphics*[trim = 0 0 0 0, width = \wid]{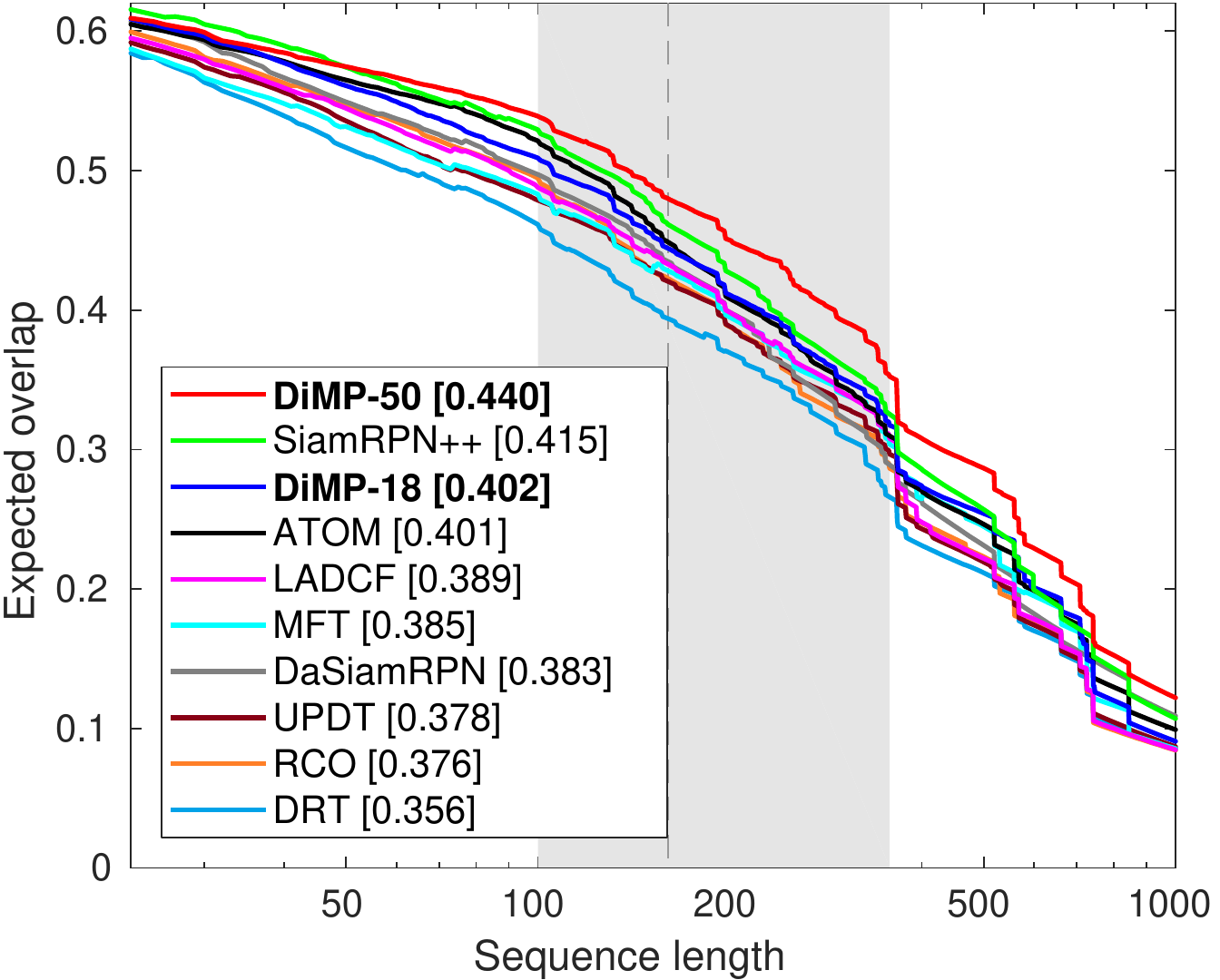}\vspace{-0mm}%
	\caption{Expected average overlap curve on the VOT2018 dataset, showing the expected overlap between tracker prediction and ground truth for different sequence lengths. The EAO measure, computed as the average of the expected average overlap over typical sequence lengths (grey region in the plot), is shown in the legend. Our approach achieves the best EAO score, outperforming the previous best approach SiamRPN++ \cite{SiamRPN++} with a relative gain of $6.3\%$ in terms of EAO. 
	}%
	\label{fig:vot_eao}%
\end{figure}

\section{Detailed Results on VOT2018}
\label{sec:VOT2018}

In this section, we provide detailed results on the VOT2018 \cite{VOT2018} dataset. The VOT protocol evaluates the expected average overlap (EAO) between the tracker predictions and the ground truth bounding boxes for different sequence lengths. The trackers are then ranked using the EAO measure, which computes the average of the  expected average overlaps over typical sequence lengths. We refer to \cite{VOT2015} for further details about the EAO computation. Figure \ref{fig:vot_eao} plots the expected average overlap for different sequence lengths on VOT2018 dataset. Our approach DiMP-50 achieves the best EAO score of $0.44$.

\section{Detailed Results on LaSOT}
\label{sec:LaSOT}

\begin{figure}[t]
	\centering%
	\newcommand{\wid}{0.9\columnwidth}%
	\includegraphics*[trim = 20 120 20 130, width = \wid]{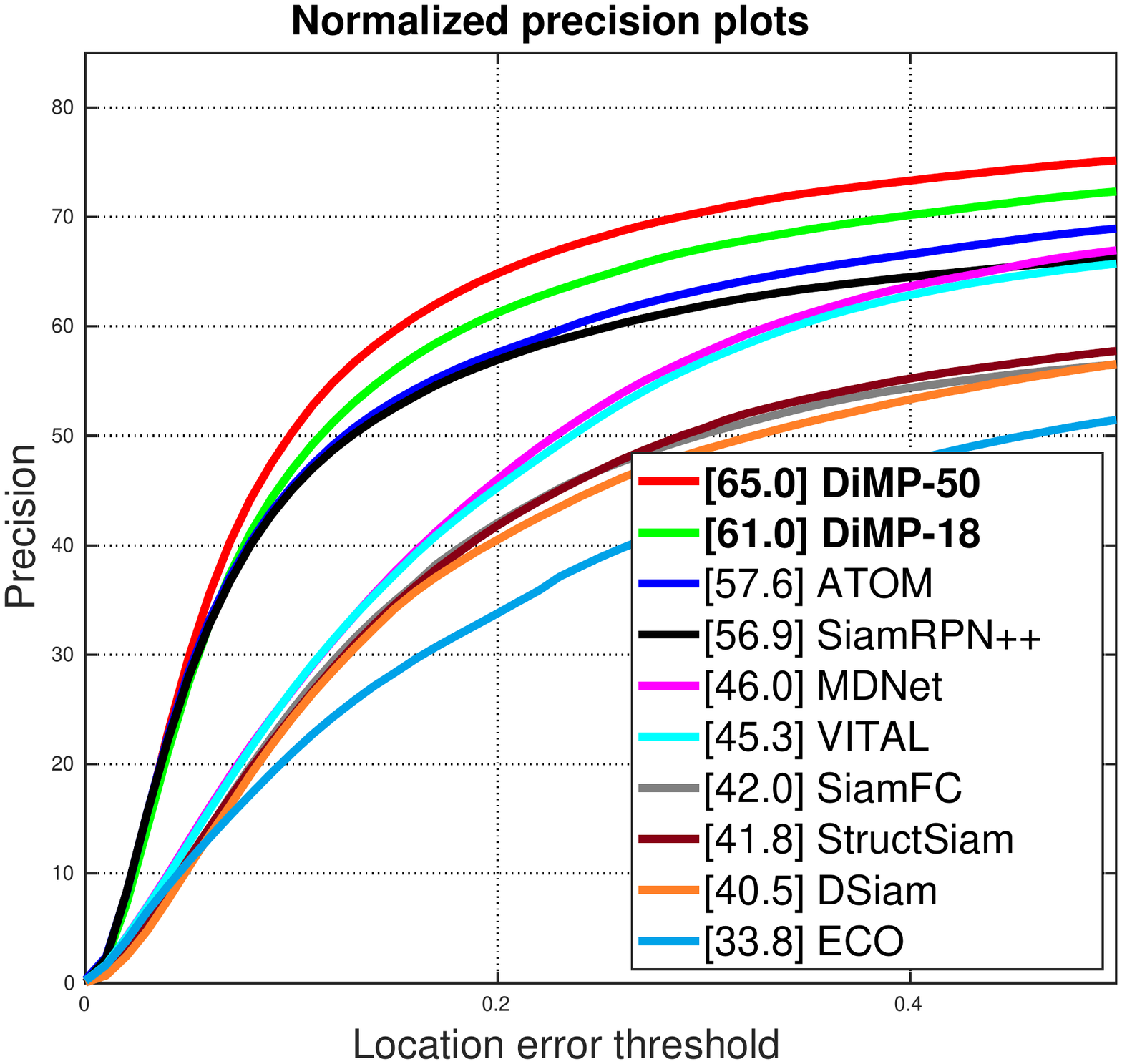}\vspace{-1.5mm}%
	\caption{Normalized precision plot on the LaSOT dataset. Both our ResNet-18 and ResNet-50 versions outperform all previous methods by significant margins.
	}%
	\label{fig:lasot_norm}%
\end{figure}

Here, we provide the normalized precision plots on the LaSOT \cite{LaSOT} dataset. These are obtained in the following manner. First, the normalized precision score $P_\text{norm}$ is computed as the percentage of frames in which the distance between the target location predicted by the tracker and the ground truth, relative to the target size, is less than a certain threshold. The normalized precision score over all the the videos are then plotted over a range of thresholds $[0, 0.5]$ to obtain the normalized precision plots. The trackers are ranked using the area under the resulting curve. Figure \ref{fig:lasot_norm} shows the normalized precision plots over all $280$ videos in the LaSOT dataset. Both our ResNet-18 (DiMP-18) and ResNet-50 (DiMP-50) versions outperform all previous methods, achieving relative gains of $5.9\%$ and $12.8\%$ over the previous best method, ATOM \cite{ATOM}.

\begin{figure*}[t]
	\newcommand{\wid}{0.33\textwidth}%
	\centering\vspace{-5mm}%
	\subfloat[NFS\label{fig:nfs}]{\includegraphics[trim = 0 0 0 0, width = \wid]{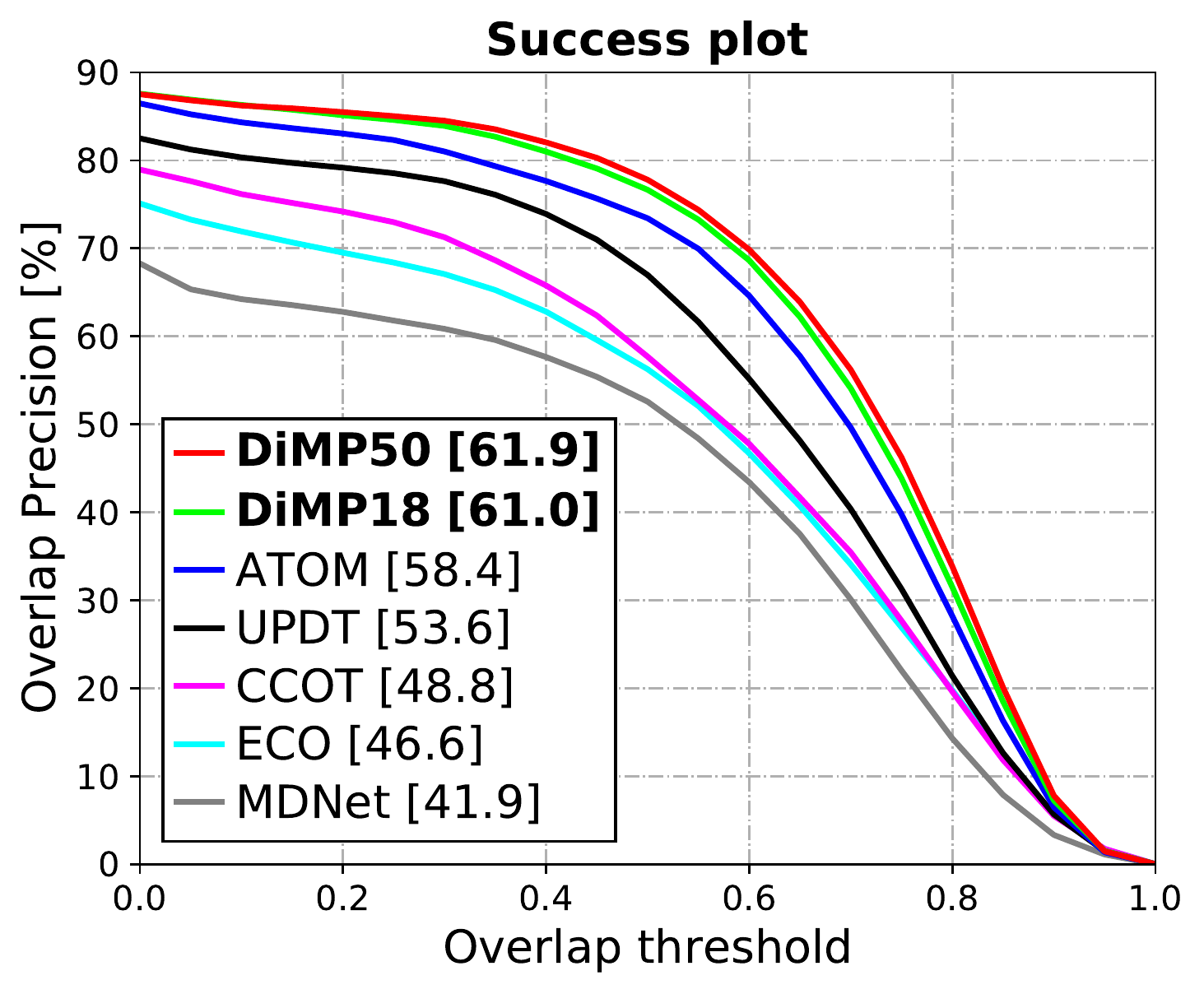}}%
	\subfloat[OTB-100\label{fig:otb}]{\includegraphics[trim = 0 0 0 0,width = \wid]{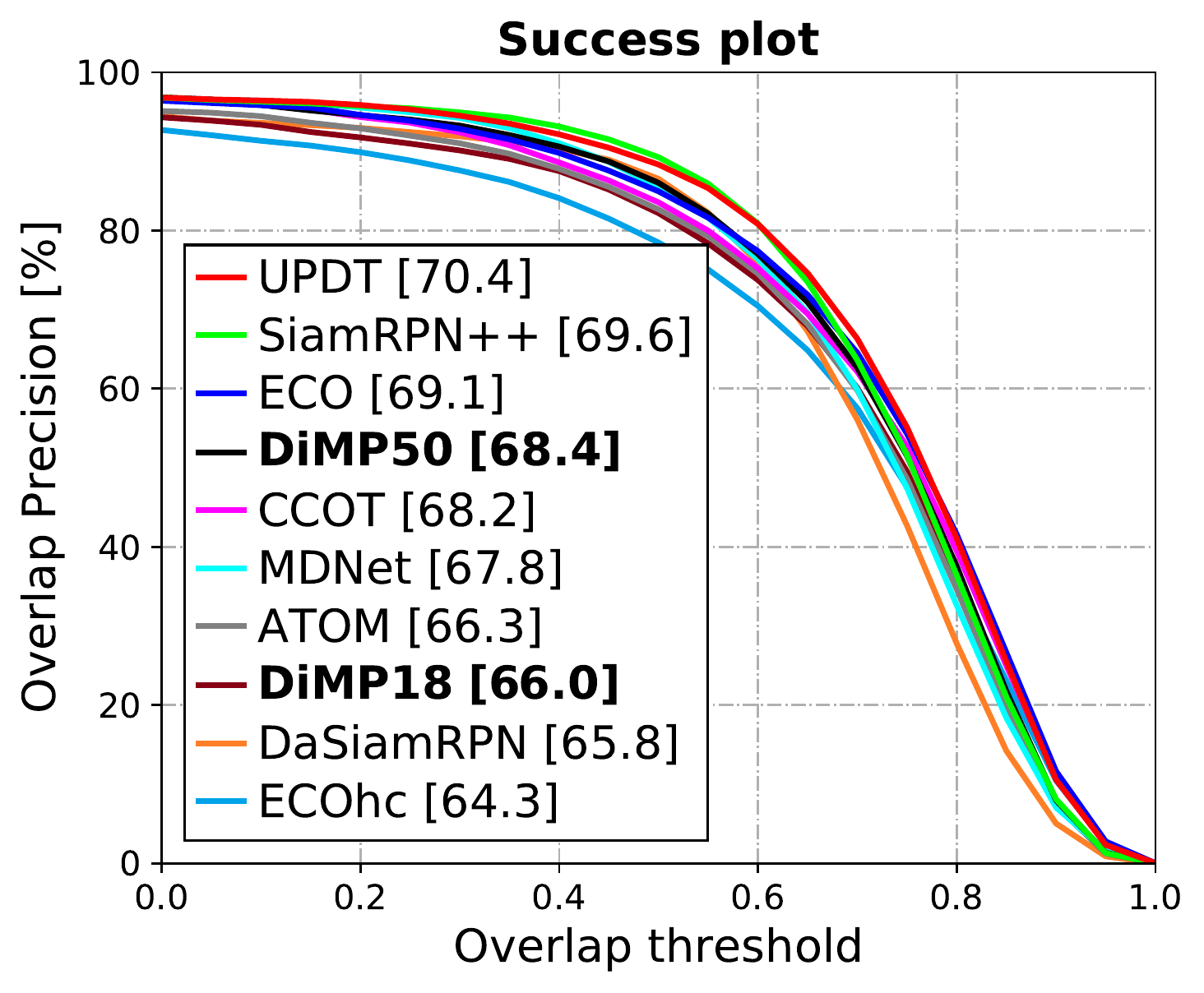}}%
	\subfloat[UAV123\label{fig:uav}]{\includegraphics[trim = 0 0 0 0,width = \wid]{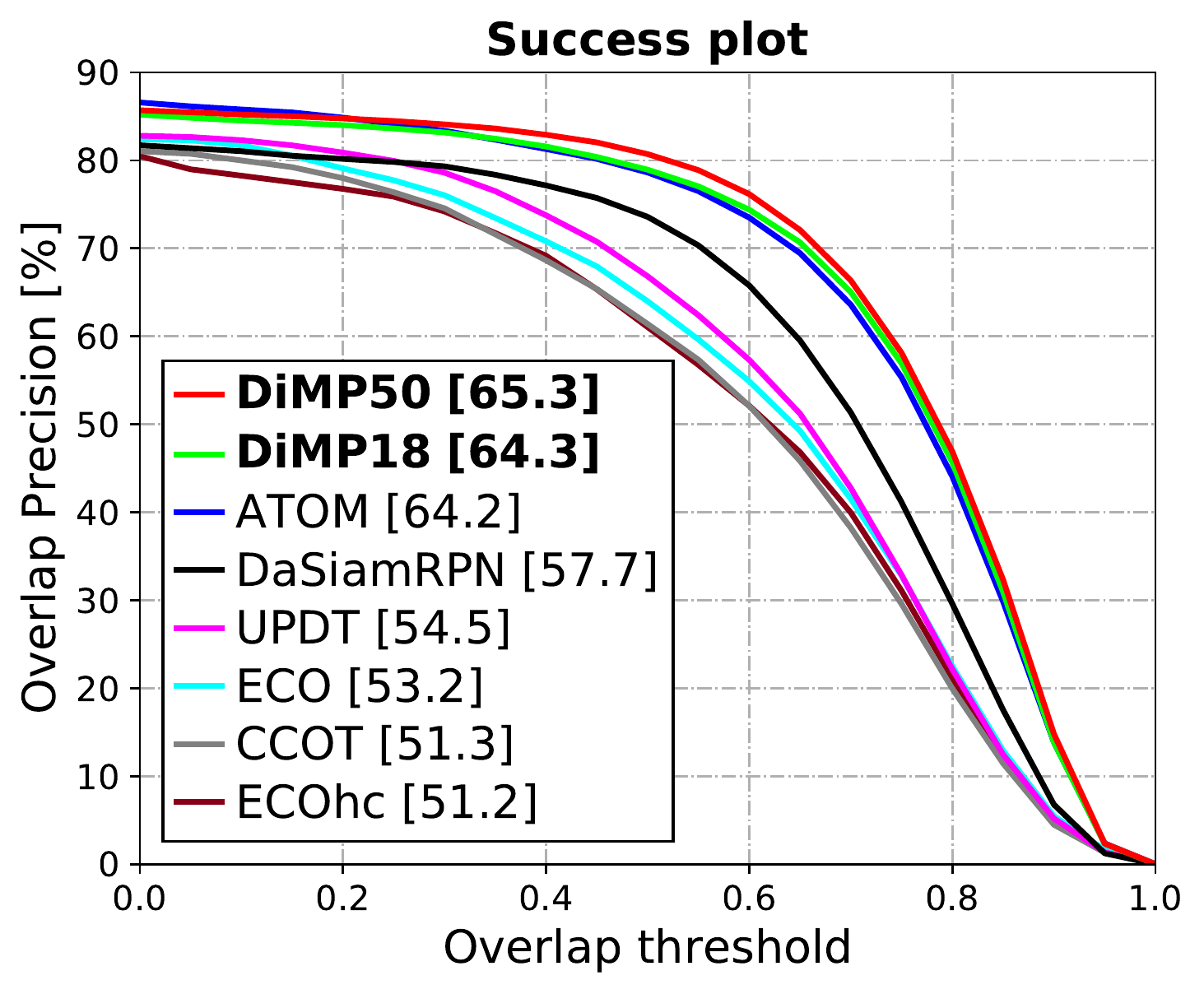}}\vspace{1mm}%
	\caption{Success plots on NFS (a), OTB-100 (b), and UAV123 (c) datasets. The area-under-the-curve (AUC) scores are shown in the legend. Our approach achieves the best scores on both the NFS and UAV123 datasets.}%
	\label{fig:success_plots}
\end{figure*}

\section{Detailed Results on NFS, OTB-100, and UAV123}
\label{sec:detailed-res}

Here, we provide detailed results on NFS \cite{NfS}, OTB-100 \cite{OTB2015}, and UAV123 \cite{UAV123} datasets. We use the overlap precision (OP) metric for evaluating the trackers. The OP score denotes the percentage of frames in a video for which the intersection-over-union (IoU) overlap between the tracker prediction and the ground truth bounding box exceeds a certain threshold. The mean OP score over all the videos in a dataset are plotted over a range of thresholds $[0,1]$ to obtain the success plot. The area under this plot provides the AUC score, which is used to rank the trackers. We refer to \cite{OTB2015} for further details. The success plots over the entire NFS, OTB-100, and UAV123 datasets are shown in figure \ref{fig:success_plots}. Our tracker using ResNet-50 backbone, denoted DiMP-50, achieves the best results on both NFS and UAV123 datasets, while obtaining results competitive with the state-of-the-art on the, now saturated, OTB-100 dataset. On the challenging NFS dataset, our approach achieves an absolute gain of $3.5\%$ AUC score over the previous best method ATOM \cite{ATOM}.

\section{Impact of Training Data}
\label{sec:less-data}
\begin{figure}[t]
	\centering%
	\newcommand{\wid}{0.95\columnwidth}%
	\includegraphics*[trim = 0 0 0 0, width = \wid]{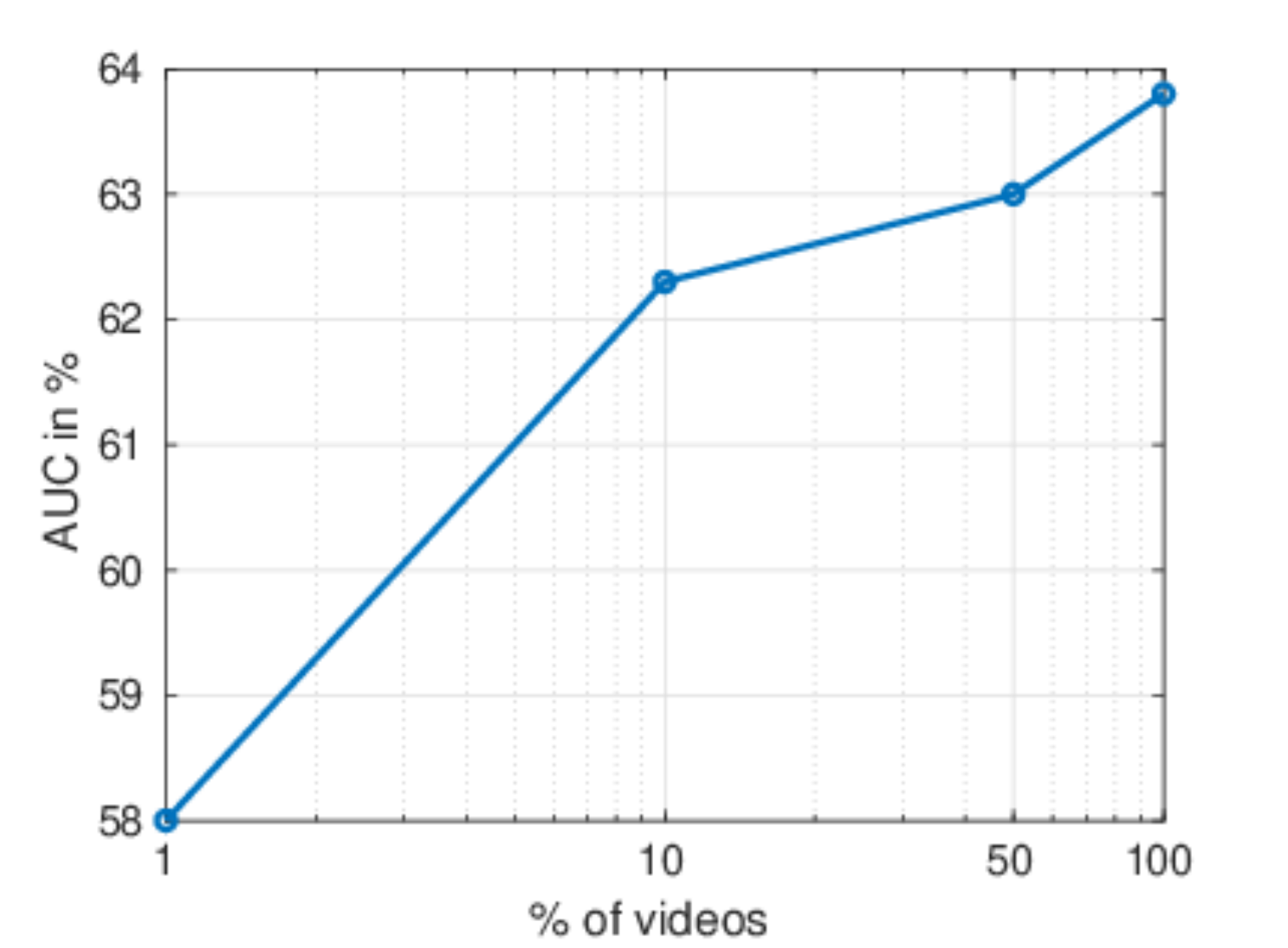}\vspace{-0mm}%
	\caption{Impact of the percentage of total videos used for offline training (log x-axis). Results are shown on the combined OTB-100, NFS, and UAV123 datasets.}%
	\label{fig:data_frac}%
\end{figure}

Here, we investigate the impact of the number of videos used for training on the tracking performance. We train different versions of our tracker using the same datasets as in the main paper, i.e.\ TrackingNet~\cite{TrackingNet}, LaSOT~\cite{LaSOT}, GOT10k~\cite{GOT10k}, and COCO~\cite{COCO}, but using only a sub-set of videos from each dataset. The results on the combined OTB-100, NFS, and UAV123 datasets are shown in figure \ref{fig:data_frac}. Observe that the performance degrades by only $1.5\%$ when the model is trained with only $10\%$ of the total videos. Even when using only $1\%$ of videos, our approach still obtains a respectable AUC score of around $58\%$.  

\section{Visualizations of learned $y_c$, $m_c$, and $v_c$}
\label{sec:label_vis}
\begin{figure*}[t]
	\newcommand{\wid}{0.25\textwidth}%
	\centering\vspace{0mm}%
	\hspace{-2mm}\includegraphics[trim =-60 -10 -50 -40, width =0.22\textwidth]{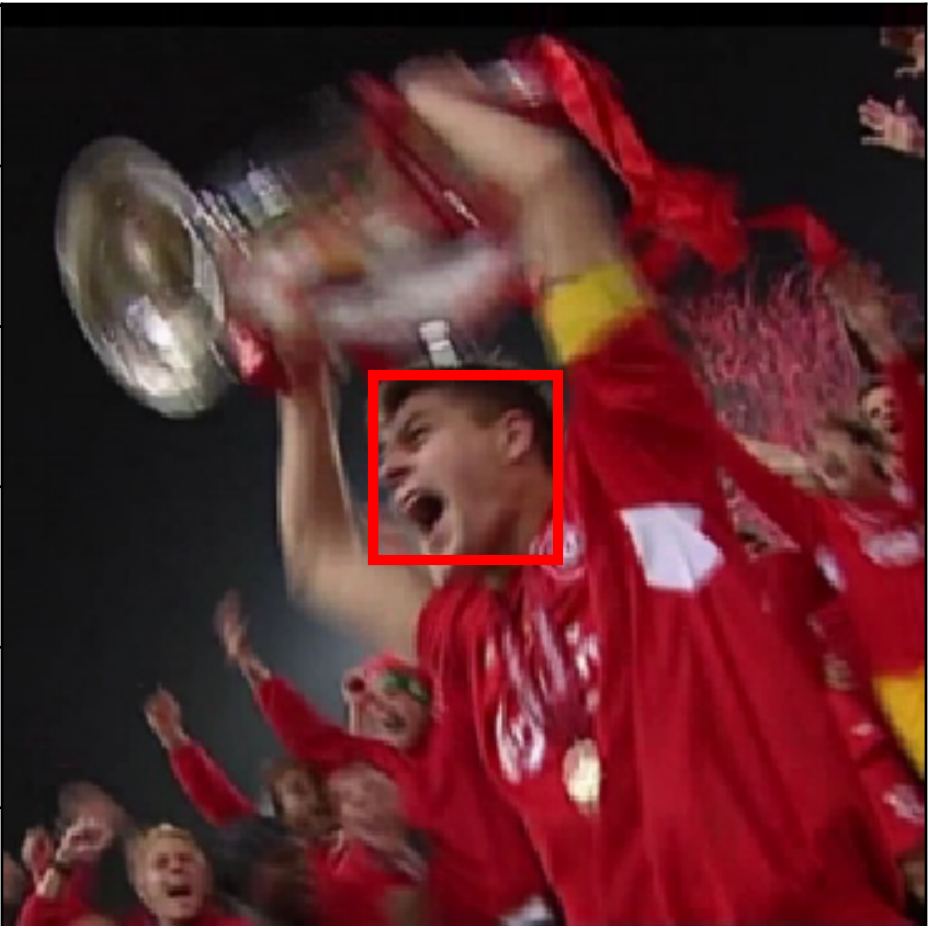}%
	\includegraphics[trim = 30 30 10 30, width =\wid]{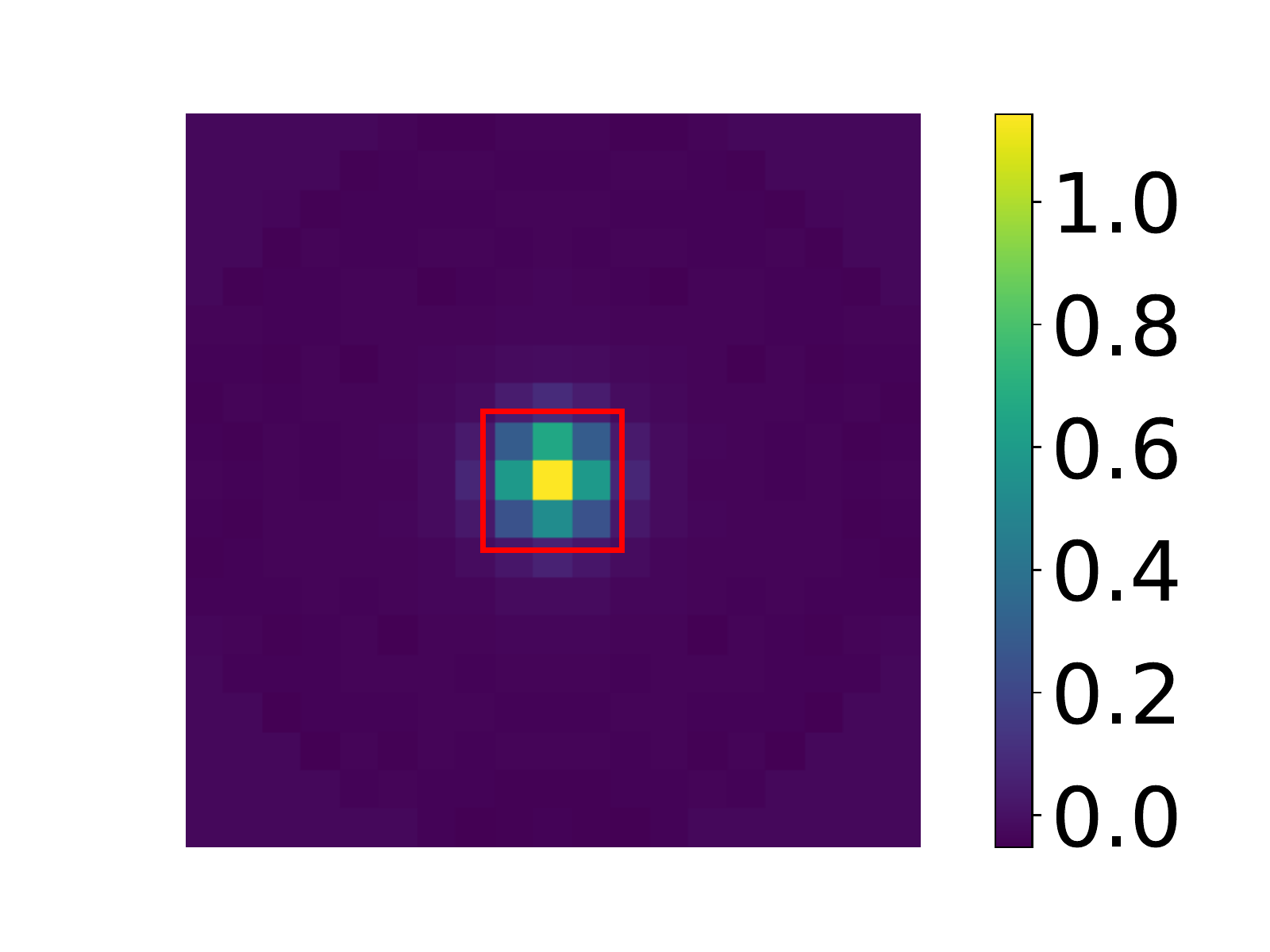}~%
	\includegraphics[trim = 30 30 10 30,width = \wid]{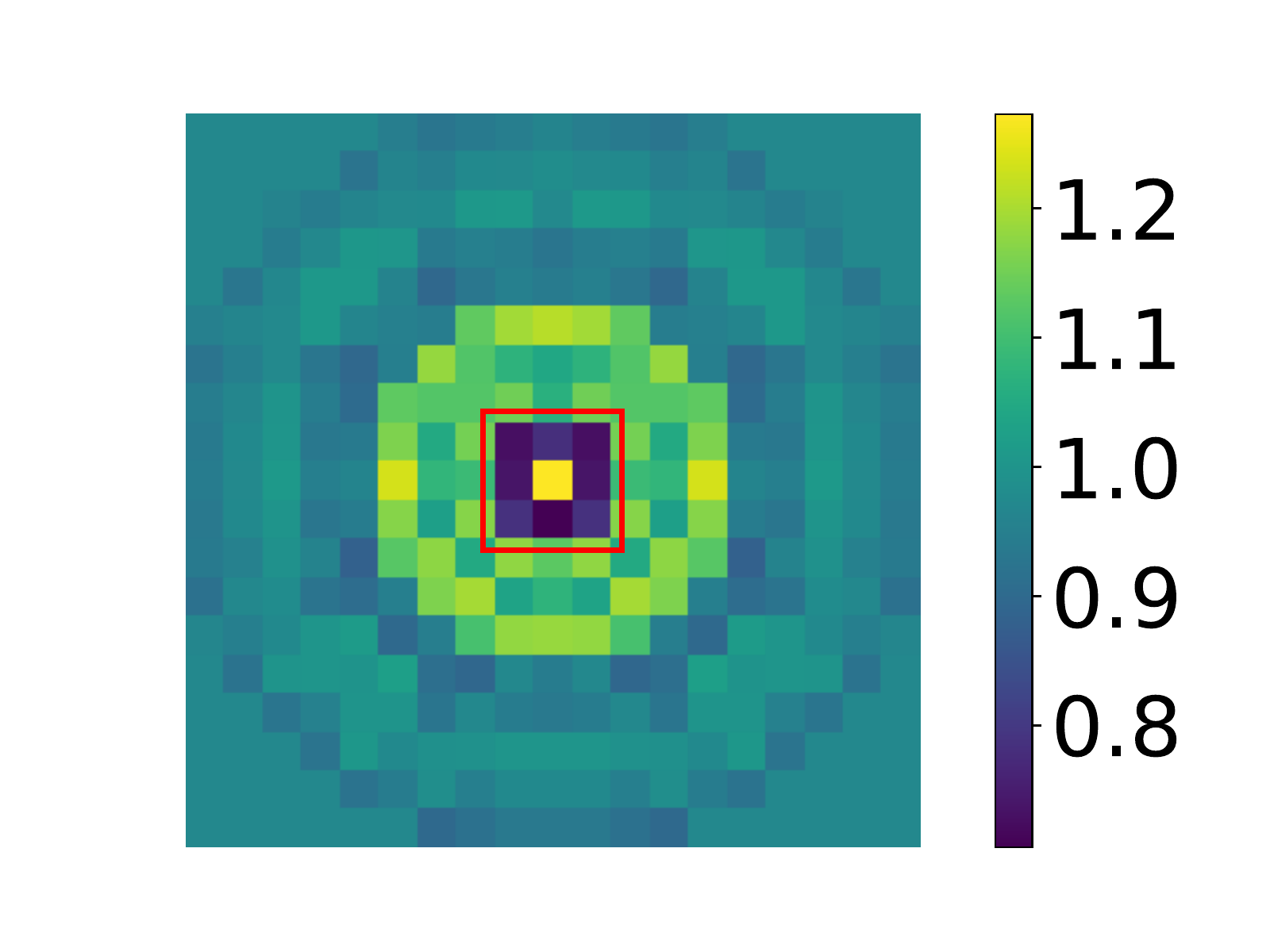}~%
	\includegraphics[trim = 30 30 30 30,width = \wid]{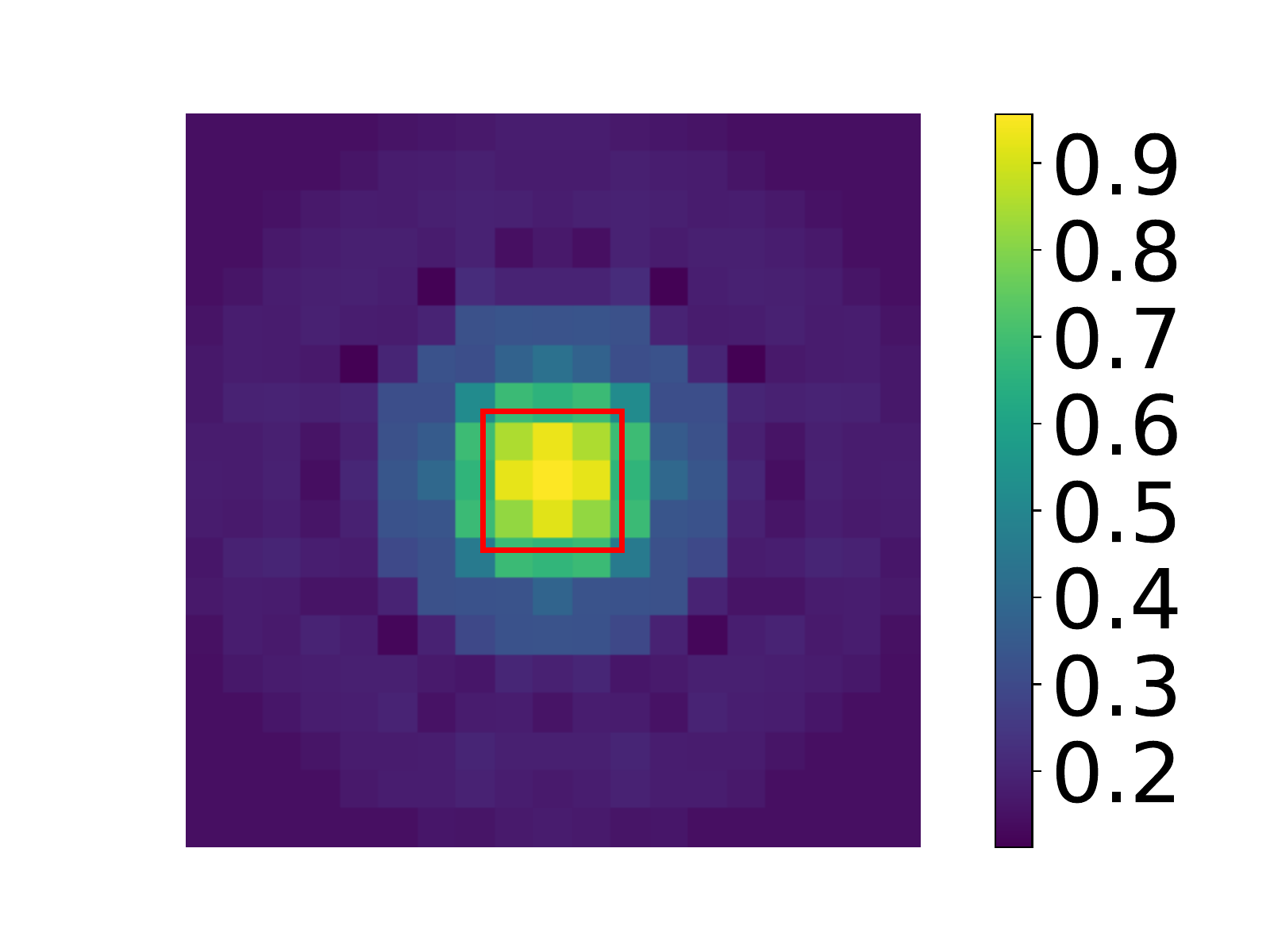}\vspace{-1mm}
	\parbox{.23\textwidth}{\centering\small Image}%
	\parbox{.24\textwidth}{\centering\small Label $y_c$}%
	\parbox{.28\textwidth}{\centering\small Spatial Weight $v_c$}%
	\parbox{.27\textwidth}{\centering\small Target Mask $m_c$}\vspace{0mm}%
	\caption{Visualization of the learned label $y_c$, spatial weight $v_c$, and target mask $m_c$. The red box denotes the target object.}%
	\label{fig:label_vis}\vspace{-4mm}%
\end{figure*}

A 2D visualization of the learned regression label ($y_c$), target mask ($m_c$), and spatial weight ($v_c$) is provided in figure \ref{fig:label_vis}. Note that each of these quantities are in fact continuous and are here sampled at the discrete feature grid points. In this example, that target (red box) is centered in the image patch. From the figure, we can see that the network learns to give the samples in the target-background transition region less weight due to their ambiguous nature.

\end{document}